\documentclass[journal]{IEEEtai}

\usepackage[colorlinks,urlcolor=blue,linkcolor=blue,citecolor=blue]{hyperref}

\usepackage{color,array}
\usepackage{multirow}
\usepackage{graphicx}

\usepackage{cite}
\usepackage{amsmath,amssymb,amsfonts}
\usepackage{algorithm}
\usepackage{algorithmic}
\usepackage{textcomp}
\usepackage{xcolor}
\usepackage{url}

\begin{document}

\title{A Survey on Masked Facial Detection Methods and Datasets for Fighting Against COVID-19 }

\author{ Bingshu Wang, Jiangbin Zheng, and C.L. Philip Chen* \IEEEmembership{Fellow, IEEE} 
\thanks{  This work was supported by the National Natural Science Foundation of China, Youth Fund, under number 62102318, in part by the Fundamental Research Funds for the Central Universities under number G2020KY05113. The work was also funded by the National Key Research and Development Program of China under number 2019YFA0706200 and 2019YFB1703600, in part by the National Natural Science Foundation of China grant under number 61702195, 61751202, U1813203, U1801262, 61751205, in part by the Science and Technology Major Project of Guangzhou under number 202007030006. (Corresponding author: C. L. Philip Chen.) }
 
\thanks{Bingshu Wang is with the School of Software, Taicang Campus, Northwestern Polytechnical University, Suzhou 215400, China (e-mail: wangbingshu@nwpu.edu.cn).}
\thanks{Jiangbin Zheng is with the School of Software, Taicang Campus, Northwestern Polytechnical University, Suzhou 215400, China (e-mail: zhengjb@nwpu.edu.cn).}
\thanks{C. L. Philip Chen is with the School of Computer Science and Engineering, South China University of Technology, Guangzhou 510641, China (e-mail: philip.chen@ieee.org).}

}

\markboth{Journal of IEEE Transactions on Artificial Intelligence, Vol. 00, No. 0, December 2021}
{First A. Author \MakeLowercase{\textit{et al.}}: Bare Demo of IEEEtai.cls for IEEE Journals of IEEE Transactions on Artificial Intelligence}

\maketitle

\begin{abstract}

Coronavirus disease 2019 (COVID-19) continues to pose a great challenge to the world since its outbreak. To fight against the disease, a series of artificial intelligence (AI) techniques are  developed and applied to real-world scenarios such as safety monitoring, disease diagnosis, infection risk assessment, lesion segmentation of COVID-19 CT scans,etc. The coronavirus epidemics have forced people wear masks to counteract the transmission of virus, which also brings difficulties to monitor large groups of people wearing masks. In this paper, we primarily focus on the AI techniques of masked facial detection and related datasets. We survey the recent advances, beginning with the descriptions of masked facial detection datasets. Thirteen available datasets are described and discussed in details. Then, the methods are roughly categorized into two classes: conventional methods and neural network-based methods. Conventional methods are usually trained by boosting algorithms with hand-crafted features, which accounts for a small proportion.  Neural network-based methods are further classified as three parts according to the number of processing stages. Representative algorithms are described in detail, coupled with some typical techniques that are described briefly.  Finally, we summarize the recent benchmarking results,  give the discussions on the limitations of datasets and methods, and expand future research directions.  To our knowledge, this is the first survey about masked facial detection methods and datasets. Hopefully our survey could provide some help to fight against epidemics. 
	
\end{abstract}

\begin{IEEEImpStatement}
	
In the era of COVID-19, many AI techniques of masked facial detection have been proposed to determine whether one wears a mask, or provide masked face regions to help non-contact temperature measurement. However, it lacks of a review about these masked facial detection methods and datasets. In this survey paper, we review recent benchmarking efforts that primarily focus on the techniques of masked face detection to combat COVID-19. We have summarized thirteen open datasets and provided their available links that would help AI researchers and engineers use them quickly. We have presented several categories of representative techniques aimed for masked facial detection. Meanwhile,  ten research directions have been identified to guide researchers for future research.  It could offer a good reference for beginners, researchers and skilled AI engineers to develop more effective and efficient systems.

\end{IEEEImpStatement}

\begin{IEEEkeywords}
Masked facial detection, Artificial intelligence, Masked face datasets, Neural networks, Broad learning system.
\end{IEEEkeywords}

\section{Introduction}

\IEEEPARstart{S}{ince} the first case was identified by COVID-19 in 2019, the coronavirus disease spread quickly and caused the outbreak all over the world in 2020  \cite{zhu2020novel,WHODeclare,roser2020coronavirus}.  According to the data released by  \cite{Johns_Hopkins}, by the end of  Dec 8, 2021, more than 267.30 millions  of humans have been identified by the COVID-19,  with more being added every day. The coronavirus disease has caused more than  5.27  millions of deaths globally. 

The COVID-19 epidemic has posed great challenge to the world. Artificial intelligence (AI) techniques are able to help people  fight against the virus in many ways  \cite{islam2021systematic,latif2020leveraging,ayumi2020application,ulhaq2020covid,niu2020decade,piccialli2021role}.  For example,  detecting masked faces  \cite{jiang2020retinamask,loey2021hybrid}, detecting COVID-19 patients  \cite{shaban2021detecting, iwendi2021classification}, assessing infection risks  \cite{guo2021assessing}, building a disease monitoring and prognosis system  \cite{kallel2021hybrid}, improving lesion segmentation of COVID-19 chest CT Scans   \cite{mahmud2021covsegnet}, etc.  Among these techniques, this survey paper primarily focuses on the techniques of masked facial detection.

Many doctors and epidemiologists have proofed that wearing a mask is an effective means to  counteract the spreading of coronavirus disease  \cite{fischer2020low,klompas2020universal,feng2020rational}.  
Detailed advice on the uses of masks was published by World Health Organization (WHO)  \cite{world2020advice}.  As a consequence, people are suggested and even required by rules or laws to wear masks when entering public places. This brings demands to monitor large groups of people wearing masks. But it is not the goal of existing face detection methods that have been embedded in monitoring  devices. To solve the problem, a series of masked facial detection methods   and   datasets have been proposed.  
 
The objective of this paper is to provide a detailed review of recent developments in the field of masked facial detection, in the hopes of providing reference or help for researchers and communities to develop more efficient and effective systems.  Current methods employ hand-crafted features and neural networks to train detection models. In this survey paper, we classify them according to the used feature and the number of processing stages. To our knowledge, this is first survey about masked face detection methods. 
    
The aims of this review paper are presented:
\begin{itemize}
	\item Describe the current open datasets of masked facial detection. Provide a detailed summary about the characteristics of datasets as well as the available links.
	\item Present a division of masked facial detection methods. For each category of techniques, representative methods are outlined and commented. 	 
	\item Perform a comparison between different methods according to the results provided by the original literatures. Give discussions about the characteristics   and limitations   of methods and provide   ten  research directions in future.   
\end{itemize}

The rest of the paper is organized as follows. Section \ref{section:StatsOfLiteratures} presents the stats of related literatures in this paper and how we surveyed literatures.    Section \ref{section:DatasetsDescription} surveys the datasets of masked face detection. Details of thirteen open datasets are outlined. Section \ref{section:MethodsDescription} gives the descriptions, main characteristics, and comparison analysis of masked facial detection methods.  Limitations of datasets and methods, and future research directions are discussed in Section \ref{section:DiscussionsAndFuture}. Conclusion is drawn in Section \ref{section:Conclusion}.

\section{The Stats and Analysis of Surveyed Literatures }
\label{section:StatsOfLiteratures} 

Since the outbreak of COVID-19 epidemic, a series of works focus on how to use AI techniques to help fight against virus. The literatures of masked facial detection are springing up around the world. Many related international conferences were held with many solutions proposed for masked facial detection  in recent two years. In this section, we shed light on the stats and analysis of state-of-the-art methods.

  \begin{figure*}
	\begin{center}
		\begin{tabular}{c}
			\includegraphics[height=7.0cm,width=17.6cm]{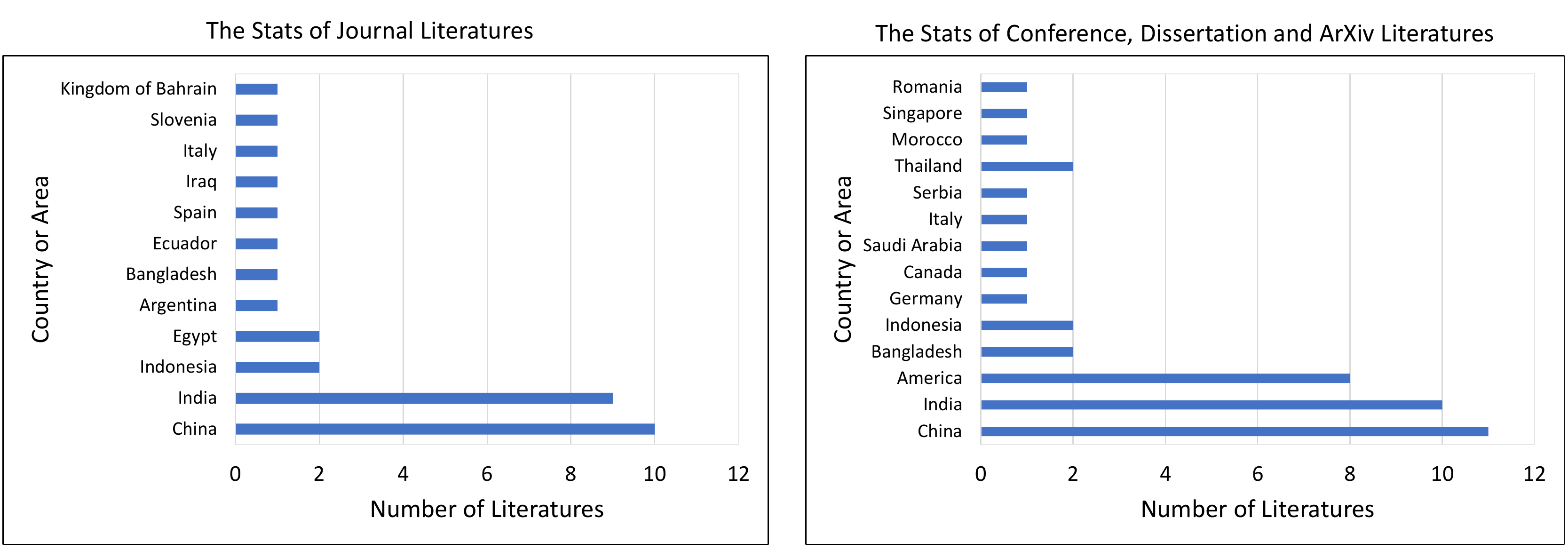}
		\end{tabular}
	\end{center}
	\caption 
	{ \label{fig:SurveyBasedOnCountries}
	 	The stats of state-of-the-art methods based on Country or Area of authors' affiliations. The literatures were surveyed by the end of September 1, 2021.     } 
\end{figure*} 

\begin{figure*}
	\begin{center}
		\begin{tabular}{c}
			\includegraphics[height=5.3cm,width=17.5cm]{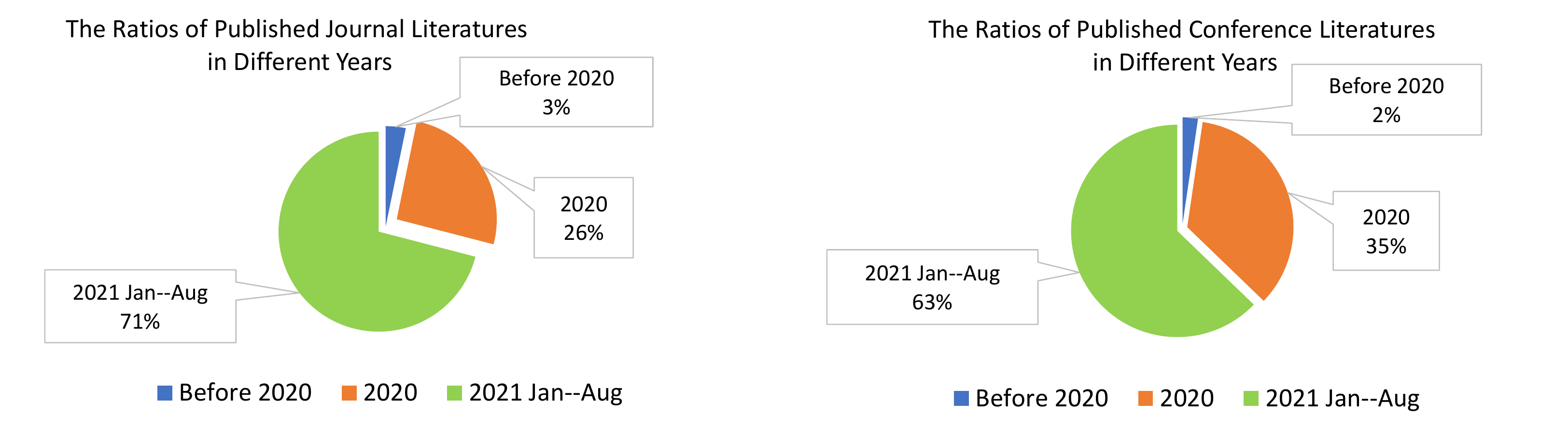}
		\end{tabular}
	\end{center}
	\caption 
	{ \label{fig:SurveyBasedOnYears}
	 	The stats of state-of-the-art methods based on published date.  Three parts are partitioned: before 2020, 2020, and 2021 Jan--Aug.    } 
\end{figure*}

\subsection{The Stats of Surveyed Literatures }

We surveyed literatures of masked facial detection by searching them in some large libraries or academic social websites such as Google Scholar, IEEEXplore, Elsevier, Springer, WebofScience, ResearchGate, etc. The searching key words are ``masked face", ``face mask", ``masked facial" with the ``document title" setting in the advanced search. 

With hundreds of items obtained, all the searched journal papers  \cite{lin2016masked,meivel2021realMatlab,meivel2021realIoT,fan2021deep,prusty2021novel,sethi2021real,adhinata2021deep,dondo2021application,nowrin2021comprehensive,yu2021face,talahua2021facial,tomas2021incorrect,hussain2021iot,mohammed2021smart,nagrath2021ssdmnv2,mercaldo2021transfer,cabani2020maskedface,chen2020efficient,qin2020identifying,mohan2021tiny,loey2021fightings,roy2020moxa,wang2021hybrid,faizah2021implementation,loey2021hybrid,omar2021automated,zhang2021novel,singh2021face,batagelj2021correctly,jiang2021real,kumar2021scaling} are selected for review due to their detailed descriptions, experiments and discussions. Some conference papers are filtered out under the conditions:  1) not written in English; 2) without experiments especially lacking of quantitative results; 3) unclear expressions or disordered organization; 4) without visual detection results shown; 5) number of images in dataset is too small, e.g., $ \leq 500$.  Specially, a few literatures utilize very similar techniques and only test algorithms  on different datasets. Only those with larger datasets and good performance are selected.   

In total,  more than 70 literatures are selected for this survey.  They cover journal papers, conference papers, dissertations, and arXivs. In this paper, we divide literatures into two classes for analysis: journal papers; conference papers. Particularly, dissertations and arXivs are assigned to conference class.   

Stats is conducted based on two ways: Country or Area of authors' affiliations; published years.   Figure \ref{fig:SurveyBasedOnCountries} outlines the number of papers for different Countries or Areas around the globe.  For the stats of Country or Area of journal papers, it is clearly concluded that most of papers are proposed by Asia and Europe. China has published the largest number of journal papers with the ratio of 32.3\%. The second largest is India with the ratio of 29.0\%. These two Asia countries contribute to more than 60\% journal papers. For the stats of  Country or Area of conference papers, China and India are still the top two countries in accordance with the number of published literatures. American ranks third. More Countries or Areas bring out conference literatures than journal literatures.  
 
For the stats of published years, Figure \ref{fig:SurveyBasedOnYears} presents a direct representation. Before 2020, very few papers are published. In 2020, the number of literatures increases significantly, with 26\% for journal class and 35\% for conference class.  Remarkably, in the first eight months of 2021, the ratio of journal literatures is much higher (71\%) than that (26\%) in 2020.  Similar comparison is shown for conference literatures. 

In summary, Asia Countries take the lead in conducting the research and publish more papers than other Countries or Areas around the world.  Since the large ratio of published literatures in 2021 Jan--Aug, it is believed that more and more papers will come forth continuously.

 \begin{figure*}
	\begin{center}
		\begin{tabular}{c}
			\includegraphics[height=8.0cm,width=18.0cm]{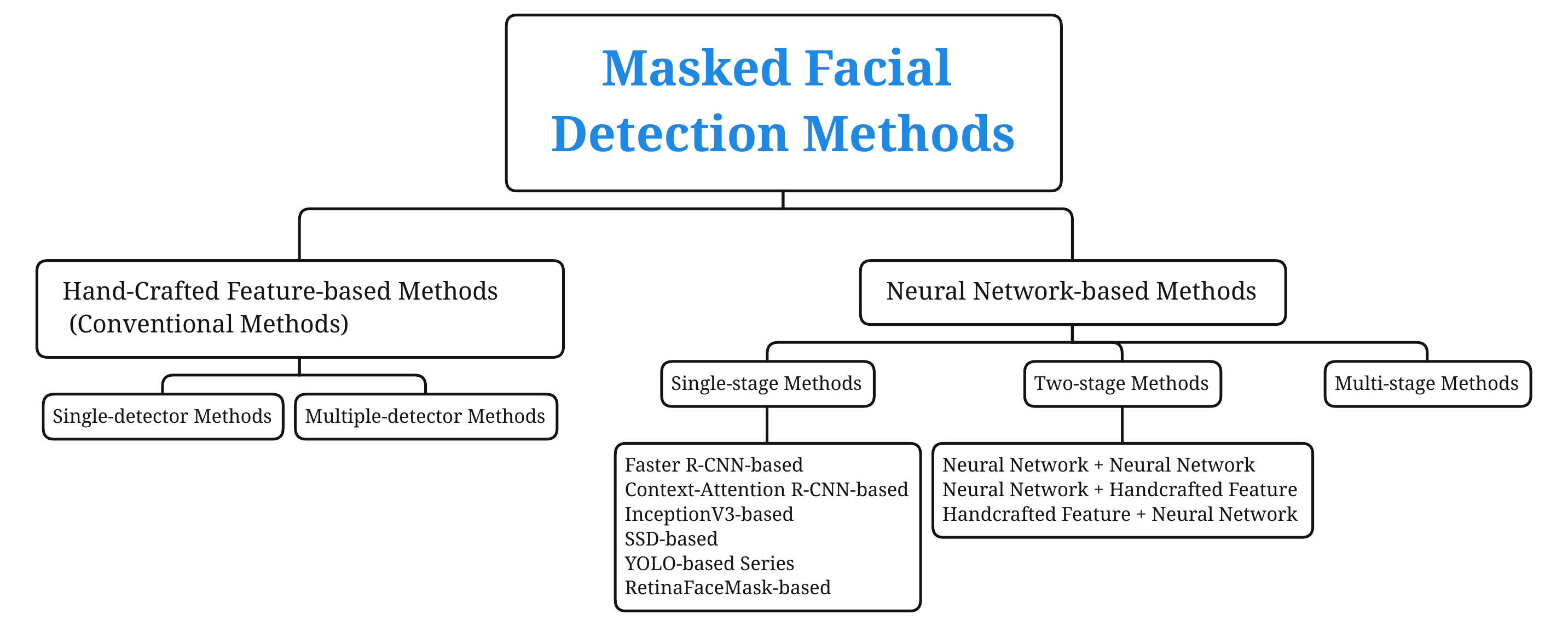}
		\end{tabular}
	\end{center}
	\caption 
	{ \label{fig:HierarchicalRepresentation}
 	A hierarchical representation of the state-of-the-art methods of masked facial detection.     } 
\end{figure*}

\subsection{The Hierarchical Representation of Surveyed Literatures}

To give a clear view of existing methods, a hierarchical representation is outlined in  Fig. \ref{fig:HierarchicalRepresentation}. According to the used features, all methods are divided into two classes: Hand-crafted Feature-based methods, Neural Network-based methods.  

Hand-crafted Feature-based methods are also usually regarded as conventional methods. They can be further classified as two categories in accordance with the number of detectors: single-detector methods and multiple-detector methods. Most detectors depend on AdaBoost algorithm. Different detectors, for example, face detector, facial mask detector, nose detector, mouth detector,  nose and mouth detector, and eye detector, are selected or combined together. Details of Hand-crafted Feature-based methods are presented in Section \ref{section:MethodsDescription}. 
 
Neural Network-based methods attract many researchers' attentions. According to the number of stages, the methods can be classified as three categories: single-stage methods, two-stage methods, and multi-stage methods. For single-stage methods, they are mainly implemented by transfer learning of object detection algorithms. For example, YOLO series methods: YOLO, YOLOv2, YOLOv3, YOLOv4, YOLOv5, and corresponding to tiny versions. For two-stage methods, they can be further divided into three kinds referring to the use of neural network: Neural Network + Neural Network, Neural Network + Hand-crafted Feature, and Hand-crafted Feature + Neural Network.   Two-stage methods consist of two parts: face region pre-detection and face region classification.  The former part is used to detect candidate facial regions, and the latter part is to classify the conditions of mask-wearing.  For multi-stage methods, they include more and complex processing steps or make use of more than one models, which means more computation costs. 

Notably, we also spend much time on the datasets of masked face detection, especially open-source datasets.  Due to their accessibility, thirteen datasets are reviewed in Section \ref{section:DatasetsDescription}.

\section{Masked facial Detection Datasets}
\label{section:DatasetsDescription} 

To monitor the conditions of wearing masks, many datasets are proposed by researchers around the globe to train detection or classification models. These models will be deployed in monitoring systems or edge-nodes.  In this section, detailed descriptions and discussions on these datasets are presented. 
 
\subsection{Description of Datasets}
 Firstly, we  present   an earlier dataset about masked face detection. 
 Ge et al \cite{ge2017detecting} proposed a large dataset called MAFA  in 2017. It was claimed to be the largest wearing mask dataset before 2017. MAFA contains 30811 images that are collected from the Internet, with 35806 masked faces. The dataset is more likely to be an occluded face dataset because it covers many mask types, for example, man-made object with single color, hand, hair, neckerchief, medical mask, etc.  The dataset is labeled with six attributes: location of face, location of eyes, location of mask, face orientation, occlusion degree, and mask type.  It considers about 60 scenes of masked faces, and provides sufficient samples. However, many occlusions are noneffective to protect people from infection risks. This dataset is more suitable for occluded face detection. Pre-processing is required to reach the goal of wearing mask detection. 

 Wang et al \cite{wang2020masked} created a Masked Face Detection Dataset(MFDD).  The dataset only concentrates on single class: masked face. It has 4342 images with a total number of 24771 masked faces. These images are captured from scenes of fighting coronavirus epidemics.  They are divided into three sets in accordance with image size \(256\times256\): equal to the size, smaller or larger than the size. The dataset can be used to train detection model to determine whether one wears a mask or not. However, it lacks of annotation information. 
 
 Cabani et al \cite{cabani2020maskedface} developed a MaskedFace-Net to generate  simulated   correct/incorrect masked faces called ``MaskedFace-Net Image Dataset(MFNID)". The framework encompasses four steps:  candidate face detection, facial landmarks detection, mask-to-face mapping, and manual image filtering. Original face images are derived from FFHQ dataset \cite{karras2019style}.  All the images have a fixed size of   \(1024\times1024\)   and they are classified as two sets: Correct Masked Face Dataset(CMFD) and Incorrect Masked Face Dataset(IMFD). The authors presented a further division for IMFD: mask only covering nose and mouth(IMFD1), mask only covering mouth and chin(IMFD2), mask only covering chin(IMFD3).  The total number of this dataset is 137016: 67193  correctly masked (49\%) and  69823 Incorrectly masked (51\%)(IMFD1, IMFD2, IMFD3). This is a very large dataset in terms of image number. For each image, facial region account for a large ratio, making face detection easy. However, MFNID only contains one type of   simulated  mask and does not provide annotations.  
 
 Roy et al \cite{roy2020moxa} searched images from the Internet  to build a  dataset namely Moxa3K.  It consists of 3000 images. The dataset gives a careful consideration for boundary conditions, for example,  if a face is covered by a handkerchief, it will be regarded as a `mask' class.  Moxa3K  includes a variety of samples such as blurred, rotated, crowded areas, and different illumination conditions.  With 9161 faces and 2015 masked faces included, all the face regions are  annotated by Pascal VOC format ``LabelImg" \cite{labelImg} and YOLO format.  Thus, it offers more choices for researchers to train their machine learning models. This setting is expected to improve the robustness of masked facial detectors.

Jiang et al \cite{jiang2021real} proposed a Properly Wearing Masked Face Detection(PWMFD) Dataset. They collected   9205 images from several available datasets such as MAFA \cite{ge2017detecting}, MFDD \cite{wang2020masked}, Wider Face \cite{yang2016wider}, and the Internet. Although several datasets have their own annotations, PWMFD dataset provides uniform annotation manually for three classes ``with\_mask'', ``without\_mask", and ``Incorrect\_mask". Specially,  facial  regions that are covered by other objects are labeled as ``without\_mask" so that  trained models are not deceived.  Face regions with nose uncovered are annotated as  ``Incorrect\_mask" class. PWMFD dataset has 7695 ``with\_mask'' faces, 10471 ``without\_mask" faces, and 366  ``Incorrect\_mask" faces.  
  
Eyiokur et al \cite{eyiokur2021computer} proposed a Unconstrained Face Mask Dataset(UFMD) by collecting images from available datasets FFHQ \cite{karras2019style}, LFW \cite{huang2014labeled}, CelebA \cite{liu2015deep}, Youtube videos and the Internet.  These publicly images allow UFMD be a complex dataset that covers  ethnicity, age, gender, indoor and outdoor scenarios. A large amount of head pose variations are also considered in UFMD, which help improve robustness of masked face detectors.  UFMD consists of 21316 images with three classes: 10618 images with masked faces,  10698 images without masks, 500 images with incorrect masks. The authors claimed that the website will be available soon.  
 
Batagelj et al \cite{batagelj2021correctly} compiled a dataset called ``Face-Mask-Label Dataset(FMLD)" by searching images from  Wider Face \cite{yang2016wider} and MAFA  \cite{ge2017detecting} datasets. Real-world conditions are considered in FMLD: head pose, illumination, and image quality.  Only when the faces are covered by nose, mouth and chin, even the occlusions are something similar to a scarf or handkerchief, they are regarded as masked face class.   Face samples are selected from  Wider Face \cite{yang2016wider} to balance the classes, which requires a small size of 40 pixels for the height and width of each face, i.e., \(min(width,height)>40\). Thus, the face region size is not small.  Incorrect masked faces are selected from those samples with nose uncovered in MAFA.   Through  inspecting samples carefully, a total number of 41934 images (63072 faces) are created in FMLD. It contains three classes of faces with labels: 32012 faces without masks, 29532 correct masked faces, 1528 incorrect masked faces.       
 
Dey et al \cite{dey2021mobilenet} created a dataset containing 4095 images that can be obtained from the available link in Table \ref{tab:OpenDatasetLink}.  Most of images have only one face. The images are selected from MFDD \cite{wang2020masked} and SMFD \cite{Prajnasb}.  Dey's dataset consists of two classes: 1930 faces without masks and  2165 faces with masks.  Head poses vary from frontal to profile. Most of scenes are simple because face regions account for large ration in the whole image.  However, annotations are not provided. 

Singh et al \cite{singh2021face} generated a custom dataset manually which includes 7500 images: 5191 training images, 1599 validation images, 710  testing  images. These images come from MAFA \cite{ge2017detecting} and Wider Face \cite{yang2016wider}.  Singh's dataset is labeled by two  classes: ``face'' and ``face\_mask'', which aims to train a model to determine whether one wears a mask or not.  The detection results can be used to analyze the crowing extent.  Bounding boxes are provided as annotations.   

Wang et al \cite{wang2021hybrid}  proposed a Wearing Mask Detection(WMD) dataset with 7084 images.  Most of the images are collected from the scenarios of combating COVID-19 in China, which allows the dataset be real-world scenarios. The dataset has a total number of 26403 masked faces: 17654 for train, 1936 for validation, and 6813 for test. It should be noted that for the test set is divided into three parts according to the  difficulty of detection task and number of masked faces in one image: DS1, DS2, DS3. Every image in DS1 has only one masked face with a relative big size. Every image in DS2 has two to four masked faces. For DS3, over five masks are included in each image and the distance from face to camera is long (\(>2m\)).  Thus, the difficulty varies from easy to difficult for the three sets. In addition, the authors also present a self-built face detection dataset which has 4054 images with 16216 faces. Coupled with WMD, these datasets can be utilized together to train models of detecting the conditions of wearing masks.
 
Moreover, there are some datasets proposed with available links such as AIZOOTech \cite{AIZOOTech}, Kaggle \cite{Kaggle}, SMFD \cite{Prajnasb}, etc.  The images of AIZOOTech \cite{AIZOOTech} dataset are from MAFA \cite{ge2017detecting} and Wider Face \cite{yang2016wider} datasets. The total number of images is 7959: 4034 masked faces and 12620 faces. Notably,  all selected images belong to scenes with medium-level difficulty. 

Kaggle \cite{Kaggle} dataset has three classes: faces without masks, correct wearing masks, incorrect wearing masks.  It consists of 853 images in total: 3232 faces with masks, 717 faces without masks,and  123 incorrect masked faces. 

SMFD dataset was proposed by Prajnasb \cite{Prajnasb} and   simulated  totally by matching masks to faces. All the original images are captured from Web. It has two categories of faces with annotations: 690 with masks and 686 without masks.  The head pose is from frontal to profile and the size of facial region is big. All these  elements  lead to a simple scene.    

In summary, detailed information for above mentioned datasets is illustrated in Table \ref{tab:OpenDataset}. The corresponding available links are also provided in Table \ref{tab:OpenDatasetLink}. All the links had been verified to be effective before May 10, 2021.

\begin{table*}[htbp]  
	\caption{   Detailed Descriptions of Some Open Datasets for Masked Facial Detection.   }  
	\begin{center}
		\centering	
		\renewcommand\arraystretch{1.25} 
		\begin{tabular}{|p{1.cm}|p{3.0cm}|p{1.1cm}|p{1.0cm}|p{1.2cm}|p{1.8cm}|p{0.8cm}|p{0.8cm}|p{0.9cm}|p{1.2cm}|p{0.6cm}| }
			\cline{1-11} 
			\textbf{ Dataset Name} & \textbf{  Main Characteristics}   & \textbf{Image Reality}    & \textbf{ Image Number}   &  \textbf{ Category}    & \textbf{ Masks Number} & \textbf{Scale }& \textbf{Head Pose} &\textbf{ Scene} & \textbf{Annotation} &  \textbf{Open} \\ \hline

			\textbf{MAFA \cite{ge2017detecting} }  &  All the images are  from the Internet. Six attributes are manually annotated for each face region.  More like occluded faces dataset.  &  Real  & 30811  &    Multiple mask types   & 35806 masked faces & Medium  Large & Various & Complex & Yes &  Yes \\ \hline

			\textbf{MFDD \cite{wang2020masked} }  &  The images are from the Internet.  Some images are collected from the scenarios of fighting against COVID-19.   &   Simulated  Real  & 4342  &    One   & 24771 masked faces & Small Medium  Large & Various & Complex & No &  Yes \\ \hline

			\textbf{MFNID \cite{cabani2020maskedface} }  &  Face images are from FFHQ. All the masks are simulated by proposed MaskedFace-Net. It includes three classes of incorrect masked faces.    &    Simulated  & 137016  &    Two   & 67193 faces with correct masks;  69823 faces with incorrect masks &  large & Frontal & Simple  & No &  Yes \\ \hline

			\textbf{Moxa3K \cite{roy2020moxa} }  &  The images are captured from Kaggle data set that are captured from Russia, Italy and China, India during the ongoing  pandemic.      &    Real  & 3000  &    Two   & 9161 faces without masks; 3015 masked faces &   Small Medium  Large & Various & Complex  & Yes &  Yes \\ \hline

			\textbf{PWMFD \cite{jiang2021real} }  & Over half of the images are collected from  WIDER Face, MAFA, RWMFD.   ``With mask" class requires  faces with nose and mouth covered.        &    Real  & 9205  &    Three   & 10471 faces without masks; 7695 correct masked faces; 366 incorrect masked faces &   Small Medium  Large & Frontal to Profile & Medium   & Yes &  Yes \\ \hline

			\textbf{UFMD \cite{eyiokur2021computer} }  & The images are captured  from FFHQ, CelebA, LFW, YouTube videos, and the Internet.  It covers ethnicity, age, gender, head pose variations.       &    Real  & 21316  &    Three   & 10698 faces without masks; 10618  correct masked faces; 500 incorrect masked faces  &    Large & Frontal to Profile & Medium   & Yes &  Soon Open \\ \hline

			\textbf{FMLD \cite{batagelj2021correctly} }  &  The images are from MAFA and  Wider Face datasets. The annotations with a list of images publicly available are provided.      &    Real  & 41934  &    Three   & 32012 faces without masks; 29532 correct masked faces; 1528 incorrect masked faces  &    Medium Large  & Various & Complex   & Yes &   Yes \\ \hline

			\textbf{Dey's Dataset \cite{dey2021mobilenet} }  & The images are real wearing masks and they come from Kaggle datasets, RMFD dataset and Bing Search.        &   Simulated  Real    & 4095  &    Two   & 2165 images with masks; 1930 images without masks &     Large  & Frontal to Profile & Simple   & No &   Yes \\ \hline 
			
			\textbf{Singh's Dataset \cite{singh2021face} }  & The dataset includes MAFA, WIDER FACE and captured images by surfing various sources.     &    Real  & 7500  &    Two   &  5191   training images; 1599 validation images; 710  testing  images &     Small  Medium  Large
			& Various & Complex   & Yes &   Yes \\ \hline 
			
			\textbf{WMD \cite{wang2021hybrid} }  & Most of the images are collected from real scenarios of fighting against CoVID-19.  It covers many long-distance scenes.         &    Real  & 7804  &    One  &  26403 masked faces  &      Small Medium  Large & Various & Complex   & Yes &   Yes \\ \hline 
			
			\textbf{AIZOO -Tech  \cite{AIZOOTech} }  & The dataset is created by modifying the wrong annotations from datasets of  WIDER Face and MAFA.        &    Real  & 7959  &    Two  &  12620 faces without masks; 4034 masked faces  &     Small Medium  Large
			& Various & Medium   & Yes &   Yes \\ \hline 
			
			\textbf{Kaggle \cite{Kaggle} }  & The images are all from the Internet for training  two-class models.       &    Real  & 853  &    Three  &  717 faces without mask; 3232 correct masked faces;  123 incorrect masked face &     Small Medium  Large & Various & Complex   & Yes &   Yes \\ \hline 
			
			\textbf{SMFD  \cite{Prajnasb} }  & All the images are webscrapped.    &   Simulated   & 1376  &  Two  &  686 faces without masks; 690 masked faces & Large &  Frontal to Profile & Simple   & Yes &  Yes \\ \hline 
			
		\end{tabular}
		\label{tab:OpenDataset}
	\end{center}	
\end{table*}

\begin{table*}[htbp]  
	\caption{ Available Websites of Open Source Datasets.  }  
	\begin{center}
		\centering	
		\renewcommand\arraystretch{1.25} 
		\begin{tabular}{|p{2.5cm}|c|p{2.5cm}|  }
			\cline{1-3} 
			\textbf{ Dataset Name} & \textbf{  Available Link }   & \textbf{Access Date}  \\ \hline
			
			\textbf{MAFA \cite{ge2017detecting} } & https://drive.google.com/drive/folders/1nbtM1n0--iZ3VVbNGhocxbnBGhMau\_OG & March 2, 2021\\ \hline 
			
			\textbf{MFDD \cite{wang2020masked} }  &  https://github.com/X-zhangyang/
			Real-World-Masked-Face-Dataset & November 26, 2020 \\ \hline 
			
			\textbf{MFNID \cite{cabani2020maskedface} } &  https://github.com/cabani/MaskedFace-Net &  February 22, 2021\\ \hline 
			
			\textbf{Moxa3K \cite{roy2020moxa} }  & https://shitty-bots-inc.github.io/MOXA/index.html & April 22, 2021\\ \hline 
			
			\textbf{PWMFD \cite{jiang2021real} } & https://github.com/ethancvaa/Properly-Wearing-Masked-Detect-Dataset & April 22, 2021 \\ \hline 
			
			\textbf{UFMD \cite{eyiokur2021computer} }  &  https://github.com/iremeyiokur/COVID-19-Preventions-Control-System &  August 30, 2021  \\ \hline 
			
			\textbf{FMLD \cite{batagelj2021correctly} } & https://github.com/borutb-fri/FMLD &  April 23, 2021\\ \hline 
			
			\textbf{Dey Dataset \cite{dey2021mobilenet} }  &https://github.com/chandrikadeb7/Face-Mask-Detection & April 23, 2021 \\ \hline 
			
			\textbf{Singh Dataset \cite{singh2021face} } & https://drive.google.com/drive/folders/1pAxEBmfYLoVtZQlBT3doxmesAO7n3ES1?usp=sharing & April 24, 2021 \\  \hline 
			
			\textbf{WMD \cite{wang2021hybrid} }   & https://github.com/BingshuCV/WMD  & April 29, 2021 \\ \hline 
			
			\textbf{AIZOO -Tech  \cite{AIZOOTech} }  & https://github.com/AIZOOTech/FaceMaskDetection & December 23, 2020 \\ \hline  
			
			\textbf{Kaggle \cite{Kaggle} }  &https://www.kaggle.com/andrewmvd/face-mask-detection   & April 22, 2021 \\ \hline 
			\textbf{SMFD \cite{Prajnasb} }  & https://github.com/prajnasb/observations & December 27, 2020 \\ \hline

		\end{tabular}
		\label{tab:OpenDatasetLink}
	\end{center}	
\end{table*}

\begin{figure*}
	\begin{center}
		\begin{tabular}{c}
			\includegraphics[height=23.5cm,width=18cm]{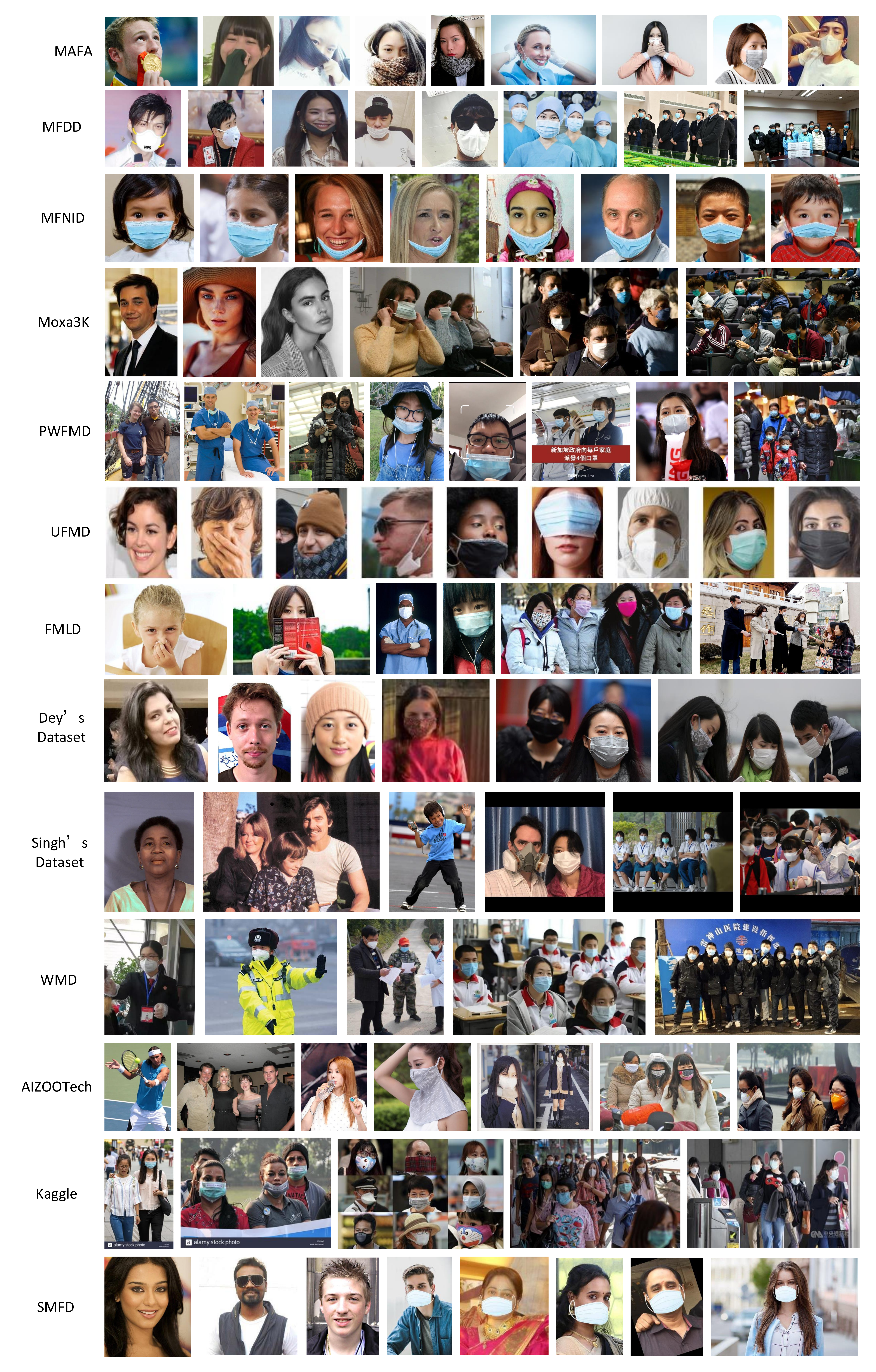}
		\end{tabular}
	\end{center}
	\caption 
	{ \label{fig:OpenDatasets}
		Some samples selected from the datasets in Table \ref{tab:OpenDataset}. These samples are the representatives of different datasets. } 
\end{figure*} 

\subsection{Discussions of Datasets}

In previous section, we elaborate on datasets and their details of characteristics. Discussions about these datasets will be presented from four parts: image sources, reality of images, classes imbalance, and existing experimental results. 

\subsubsection{Image Sources}
Almost all the datasets are created by collecting images from the Internet. A typical representative is MAFA \cite{ge2017detecting}, which is proposed as an earlier work. 
Most of images in MFDD \cite{wang2020masked}, WMD \cite{wang2021hybrid}, Kaggle \cite{Kaggle}, SMFD \cite{Prajnasb}  are built  through Internet search .

 Some faces without masks are from some face datasets such as FFHQ  \cite{karras2019style} and Wider Face \cite{yang2016wider}.   FFHQ is widely used in MFNID \cite{cabani2020maskedface} and UFMD \cite{eyiokur2021computer}.

The masked face dataset MAFA \cite{ge2017detecting} and face dataset Wider Face  \cite{yang2016wider} are  widely employed to create new masked face detection datasets such as PWMFD \cite{jiang2021real},  FMLD \cite{batagelj2021correctly},  Singh's Dataset\cite{singh2021face}, and  AIZOOTech \cite{AIZOOTech}.  This can give a good explanation about the high similarity between Singh's Dataset \cite{singh2021face} and  AIZOOTech  \cite{AIZOOTech}.  

Some datasets like PWMFD \cite{jiang2021real}  and  Dey's Dataset \cite{dey2021mobilenet} are the combinations of several existing datasets. In realistic applications, combination of multiple datasets is an alternative way to build up a required dataset quickly. Thus, it is suggested for researchers to use this way to create their own datasets. Meanwhile, capturing a variety of images from the Web is beneficial   to enrich the varieties of datasets.

\subsubsection{Reality of Images} 

It's also notable from Table \ref{tab:OpenDataset} that nine of thirteen datasets are constructed by real-world images.  MFDD \cite{wang2020masked} and  Dey's dataset \cite{dey2021mobilenet}  include both real and  simulated images.  For MFNID \cite{cabani2020maskedface}  and  SMFD \cite{Prajnasb} datasets, the masked faces are created entirely by simulating images. Some samples are given in Fig. \ref{fig:OpenDatasets}. Only one type of mask is used to synthesize masked faces in MFNID or SMFD. 

Creating  simulated samples requires a  mask-to-face mapping technique. Large size of faces are always selected to synthesize masked faces because their landmarks can be located well, which helps generate  proper  samples.  However, for small size of faces, it is hard to realize a good mapping due to the inaccurate landmarks and head pose variations. In addition, the number of mask types is inadequate. These factors allow masked face detection to be a simple problem.  This has been verified by the method \cite{wang2021hybrid}, which achieves an accuracy of 99.9\% for incorrect masked faces on 4500 images randomly selected from MFNID  \cite{cabani2020maskedface}.  

In people's daily life, there are diverse mask types. It is not easy to collect enough images with a variety of masked faces.  In this case,  synthesizing samples can be regarded as a good choice to address this issue \cite{ZamhownWear-a-mask}. It illustrates that  real mask looks more natural than the simulated  masks. More details of synthesizing images are provided in supplementary materials.    Another method of  converting face dataset to masked dataset can be found in \cite{anwar2020masked}. How to generate more natural masked faces is an interesting research in future.

\begin{table}[htbp]  
	\caption{  The Numbers and Ratios of Different Classes for Several Datasets.  }  
	\begin{center}
		\centering	
		\renewcommand\arraystretch{1.25} 
		\begin{tabular}{|c|p{1.2cm}|p{1.2cm}|p{1.2cm}|  }
			\cline{1-4} 
			\textbf{ Dataset Name}     & \textbf{ Face } & \textbf{ Correct Face\_mask } &\textbf{ Incorrect Face\_mask   }   \\ \hline
			
			\textbf{PWMFD \cite{jiang2021real} } &  10471 (56.50\%)  &  7695 (41.52\%)  & 366 (1.97\%)    \\ \hline 	
			
			\textbf{UFMD \cite{eyiokur2021computer} }  & 10698 (49.04\%)  & 10618 (48.67\%) & 500 (2.29\%)   \\ \hline 
			
			\textbf{FMLD \cite{batagelj2021correctly} } & 32012 (50.75\%) & 29532 (46.82\%) & 1528 (2.42\%)     \\ \hline 
			
			\textbf{Kaggle \cite{Kaggle} }  & 717 (17.61\%) & 3232 (79.37\%) & 123 (3.02\%)    \\ \hline 
			
		\end{tabular}
		\label{tab:ClassesBalance}
	\end{center}	
\end{table} 

\subsubsection{Classes Imbalance}  

\begin{figure}[h]
	\begin{center}
		\begin{tabular}{c}
			\includegraphics[height=4.5cm,width=8.0cm]{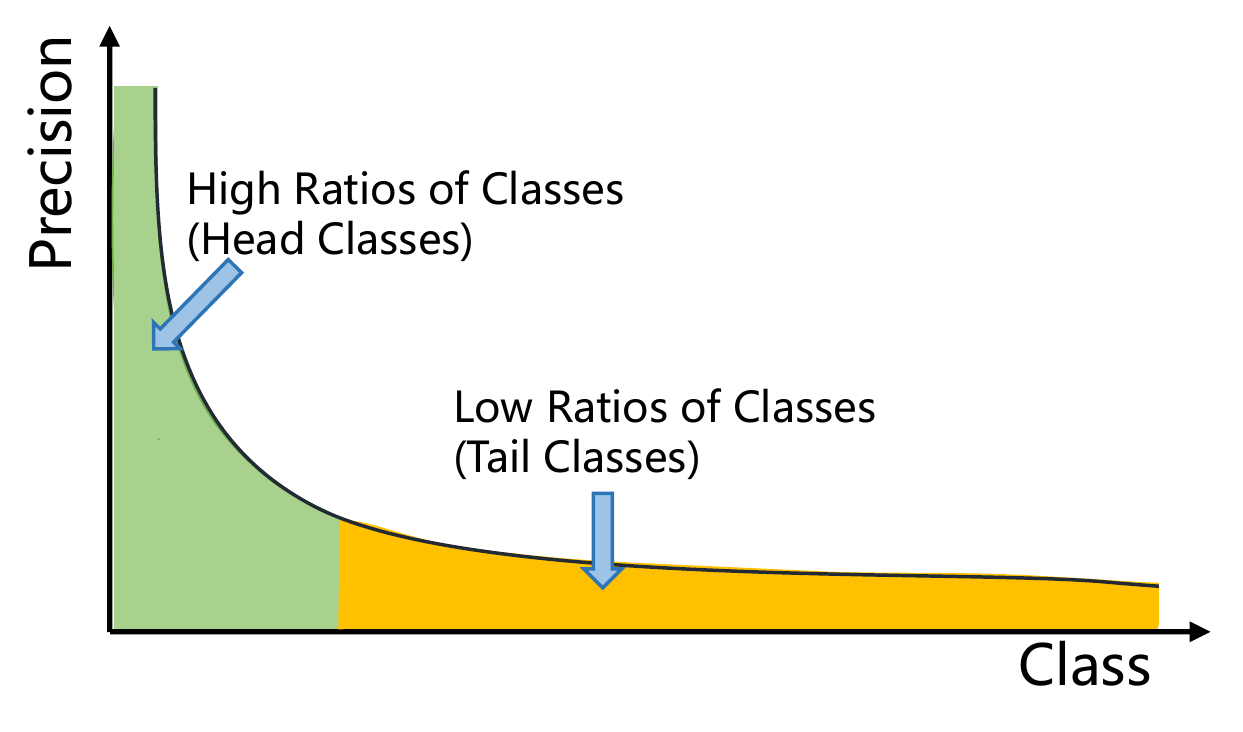}
		\end{tabular}
	\end{center}
	\caption 
	{ \label{fig:ClassImbalance}
		A distribution example of detection precisions for head classes and tail classes.     } 
\end{figure}

It's pretty clear that classes imbalance problem exists in the field of multiple categories of object detection. Table \ref{tab:ClassesBalance} sheds light on the problem for datasets PWMFD  \cite{jiang2021real}, UFMD \cite{eyiokur2021computer}, FMLD  \cite{batagelj2021correctly}, Kaggle  \cite{Kaggle}.  High ratios of classes are denoted as ``head classes", and low ratios of classes are denoted as ``tail classes".  Obviously, the ratios of incorrect face\_mask  in Table \ref{tab:ClassesBalance} are smaller than 3.1\%.  It implies that class distribution is extremely imbalanced. If a dataset with classes imbalance is used to train a  model, it will easily lead to erroneous detections.  The reason is that head classes can be learned well while tail classes are not learned well, as shown in Fig. \ref{fig:ClassImbalance}. 

How to solve the problem? Actually, it is not easy to obtain the incorrect or improperly masked faces.  Two ways are suggested to solve the problem. One way is to collect images as many as possible from available datasets.  The other way is to simulate images like MFNID  \cite{cabani2020maskedface}. 

\subsubsection{Existing Experimental Results}  
 
Table \ref{tab:ResultsOnDatasets} shows original results of some methods on their own datasets.  Herein, we firstly give some common evaluation metrics:  \(Recall\), \(Precision\), \(F1\), \(Accuracy\), \(AP\), and \(mAP\).  They are defined as follows.
\begin{equation}
\label{eq:recall}
Recall=\frac{TP}{TP+FN}
\end{equation}
\begin{equation}
\label{eq:precision}
Precision=\frac{TP}{TP+FP}
\end{equation}
\begin{equation}
\label{eq:F1}
F1=2*\frac{Recall*Precision}{Recall+Precision}
\end{equation}
\begin{equation}
\label{eq:Accuracy}
Accuracy=\frac{TP+TN}{TP+TN+FP+FN}
\end{equation}
where \(TP\) represents the true positives, \(TN\)  represents the true negatives, \(FP\)  represents false positives, and \(FN\)  represents false negatives. \(Accuracy\) represents the whole detection rate. 

\begin{equation}
\label{eq:IoU}
 IoU=\frac{|P\bigcap G|}{|P\bigcup G|}
\end{equation}
where  IoU means the overlap between predicted box \(P\) and ground truth box  \(G\).  The term  \(\bigcap\) is defined as intersection,  and \(\bigcup\) is defined as union between two boxes.  

The Average Precision (AP) is defined in   Eq. (\ref{eq:AP})  to evaluate the performance of object detection methods. It is calculated by finding the area under the \(Precision-Recall\) curve. 
\begin{equation}
\label{eq:AP}
AP_{class}=\int_{0}^{1}P(r)dr
\end{equation}
where \(class\) represents the object classes such as ``face", ``masked face", and ``incorrect masked face", etc. \(mAP\) is the mean Average Precision, as shown in  Eq. (\ref{eq:mAP}) 
\begin{equation}
\label{eq:mAP}
mAP= \frac{1}{n}\sum_{k=1}^{k=n}AP_{k}
\end{equation}

It can be clearly concluded from the Table \ref{tab:ResultsOnDatasets} that methods  \cite{eyiokur2021computer,dey2021mobilenet} achieve higher \(Accuracy\) and \(mAP\) values on the given datasets.  This also verifies the description in Table \ref{tab:OpenDataset} that scenes of UFMD and Dey's datasets are simple. The method \cite{batagelj2021correctly} largely gets benefit from two deep neural networks: RetinaFace and ResNet-152.  Other datasets such as MAFA, Moxa3K, PWMFD, Singh's, WMD are challenging in terms of quantitative results.  As a consequence, these results can be treated as benchmarks for future comparison.

 \begin{table*}[htbp]  
	\caption{The Results of Original Methods on Their Own Available Datasets.   }  
	\begin{center}
		\centering	
		\renewcommand\arraystretch{1.25} 
		\begin{tabular}{|p{3.0cm}|p{3.5cm}|p{2.2cm}|p{6.5cm}|  }
			\cline{1-4} 
			\textbf{ Literatures}     & \textbf{ Methods or Networks } & \textbf{ Datasets } &\textbf{ Results }   \\ \hline
			
			\textbf{Ge et al \cite{ge2017detecting} } &  LLE-CNNs  &  MAFA  & AP=76.4\%   \\ \hline 	
			
			\textbf{Roy et al \cite{roy2020moxa} } & SSD, Faster R-CNN,  YOLOv3, YOLOv3Tiny &  Moxa3K  &  SSD mAP=46.52\%, Faster R-CNN  mAP=60.5\%, YOLOv3 mAP=63.99\%, YOLOv3Tiny mAP=56.57\%  \\ \hline
			
			\textbf{Jiang et al \cite{jiang2021real} } &  Squeeze and Excitation-YOLOv3  &  PWMFD  & Image size \(608\times608\): AP=73.7\%  \\ \hline
			
			\textbf{Eyiokur et al \cite{eyiokur2021computer} }  & InceptionV3, ResNet-50, MobileNetV2, EfficientNet-b3 & UFMD & Three classes Accuracy: InceptionV3 98.28\%, ResNet-50 95.44\%,MobileNetV2 98.10\%,EfficientNet-b3 98.00\%  \\ \hline 
			
			\textbf{Batageli et al \cite{batagelj2021correctly} } & RetinaFace, ResNet152 & FMLD & mAP=90.75 \(\pm\) 0.99     \\ \hline 
			
			\textbf{Dey et al \cite{dey2021mobilenet} }  & MobileNetV2 & Dey's Dataset & IDS1 700 real images, Accuracy=93\%, IDS2 276 simulated images, Accuracy=100\%   \\ \hline 
			
		   \textbf{Singh et al \cite{singh2021face} }  & YOLOv3, Faster R-CNN & Singh's Dataset & YOLOv3 AP=55\%, Faster R-CNN AP=62\%    \\ \hline 
			
		    \textbf{Wang et al \cite{wang2021hybrid} }  & Faster R-CNN, BLS & WMD &  Recall=93.54\%, Precision=94.84\%, F1=94.19\%    \\ \hline 
		    
		\end{tabular}
		\label{tab:ResultsOnDatasets}
	\end{center}	
\end{table*}

\begin{figure*}
	\begin{center}
		\begin{tabular}{c}
			\includegraphics[height=3.4cm,width=17.6cm]{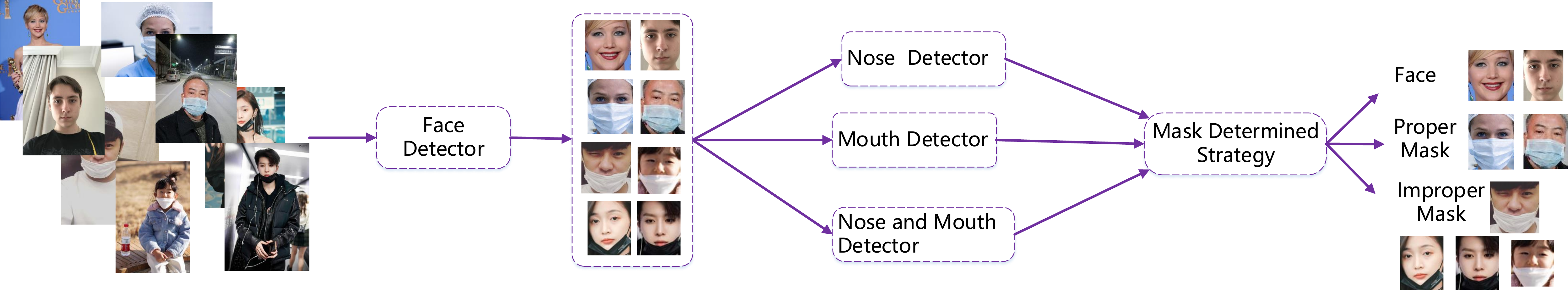}
		\end{tabular}
	\end{center}
	\caption 
	{ \label{fig:ConventionalMethods}
		A flowchart for some conventional methods aiming to identify wearing mask conditions. Several detectors are trained by self-built datasets or provided by    Open Source Computer Vision Library(OpenCV)  with its link: ``https://opencv.org/")}.    
\end{figure*}

\section{Masked Facial Detection Methods}
\label{section:MethodsDescription} 

In this section, we primarily focus on masked facial detection methods.   According to the used features, the methods can be divided into hand-crafted feature-based methods and neural network-based  methods. Hand-crafted feature-based methods are regarded as conventional methods. Specially, neural network-based methods are sprouting up and they have achieved impressive and excellent results. Considering the high proportion of neural network-based methods, we classify them as three parts based on processing stages: single-stage methods, two-stage methods, and multi-stage methods.  Detailed descriptions are given as follows.

\subsection{Conventional Methods}

Conventional face detection methods have been invested very well in past decades \cite{zafeiriou2015survey,viola2001rapid,viola2004robust,nieto2015system}. A face detector proposed by Viola and Jones \cite{viola2001rapid,viola2004robust}  is trained by 
AdaBoost algorithm, which is the basis for face detection. Common hand-crafted  features include haar-like \cite{viola2001rapid}, Local Binary Pattern(LBP) \cite{ojala1996comparative}, and Histogram  of Orientation(HOG) \cite{dalal2005histograms},etc.

In this section, we mainly focus on masked face detection using conventional methods.  Some published literatures recently are usually designed by hand-crafted features and boosting learning algorithms \cite{dewantara2020detecting,petrovic2020iot,nieto2015system,arif2021evaluation,fang2021design,he2021mask,nowrin2021comprehensive}.   Most of conventional methods for masked face detection are  based on the observation that if one wears a mask well, the nose or mouth cannot be detected, and vice versa. One typical flowchart for conventional methods is shown in Fig. \ref{fig:ConventionalMethods}.    One or several detectors are trained by self-built datasets or provided by OpenCV. Mask determined strategy is exploited to judge the mask-wearing conditions. According to the number of detectors, conventional methods can be divided into two parts: Single-detector Methods and Multiple-detector Methods.  

 \textbf{Single-detector Methods:} 
Dewantara et al \cite{dewantara2020detecting} exploited to train a nose and mouth classifier to detect multi-pose masked faces. The authors create a dataset of nose and mouth. Haar-like, LBP, HOG features are exploited for training models, respectively. If nose and mouth is not detected, the candidate facial region will be labeled ``masked”. Otherwise, it will be labeled ``No mask". It is reported that the trained classifier of nose and mouth  achieves an accuracy of 86.9\% using haar-like features, outperforming LBP and HOG. Obviously, there is further space to improve accuracy.

 \textbf{Multiple-detector Methods:}  
Petrovic et al \cite{petrovic2020iot} developed an indoor safety IoT system which adopts multiple AdaBoost cadcade-classifiers. These classifiers are provided by OpenCV to detect frontal face, nose, and mouth, respectively.  For a candidate face region, if no mouth and no nose are detected, it will be regarded as wearing a mask properly.  If nose is detected, it will be labeled as ``improper mask". If mouth is detected, it will be labeled as ``no mask".  This approach may work well in the access control system by OpenCV classifiers. However, it depends on OpenCV classifiers too much, and it does not provide details about accuracy.      

\begin{figure}
	\begin{center}
		\begin{tabular}{c}
			\includegraphics[height=2.4cm,width=8.6cm]{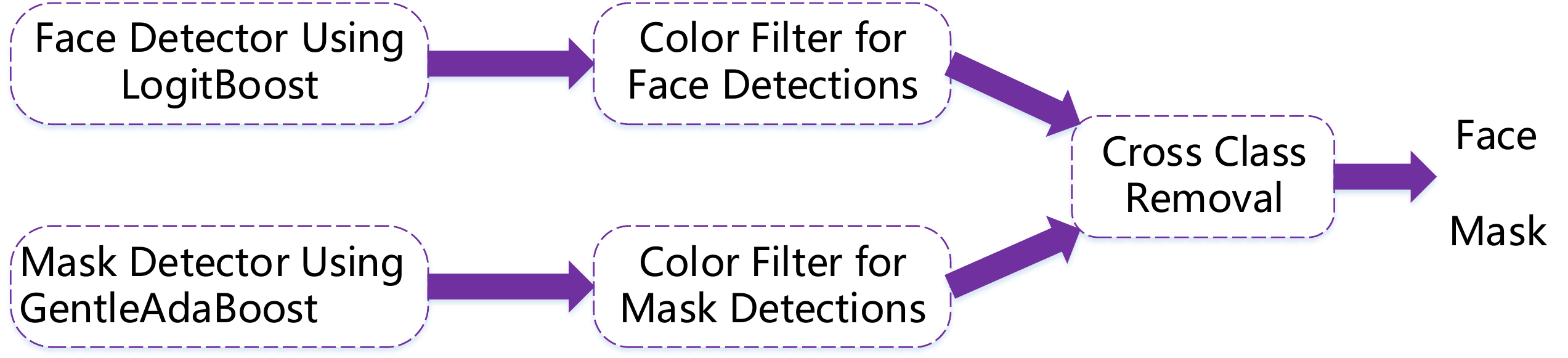}
		\end{tabular}
	\end{center}
	\caption 
	{ \label{fig:TwoAdaBoostClassifiers}
		Two AdaBoost Classifiers are used to detect faces and masks.   } 
\end{figure}

Unlike methods \cite{petrovic2020iot}, Nieto-Rodriguez et al \cite{nieto2015system} used two AdaBoost detectors to implement surgical mask detection.   One detector is trained by LogitBoost for face detection, and the other is trained by GentleAdaBoost for mask detection.  Then, two color filters in the HSV color space are employed to eliminate false positives. Considering the overlapping regions, cross class removal strategy is designed to keep the region with higher confidence. The method is easy to implement and it achieves   an accuracy of 95\%   on 496 faces and 181 masks. The process is illustrated in Fig. \ref{fig:TwoAdaBoostClassifiers}

Fang et al \cite{fang2021design} developed a real-time system of masked facial detection  that uses haar-like features for face detection and mouth detection, respectively. Similar with \cite{petrovic2020iot}, face region is firstly located, then mouth detection is used to determine the mask-wearing conditions. The designed algorithm is claimed to run on PYNQ-Z2 SoC platform with 0.13s response of facial mask detection and 96.5\% accuracy on given dataset. 

In addition, Tengjiao He \cite{he2021mask} employed skin color and eye detection to reach the goal of wearing mask detection.  The first step is to locate face region using ellipse skin model and geometric relationship between eyes and other facial parts.  Then, the coverage of skin color in the bottom half of facial region is calculated to judge  mask-wearing conditions. However, this method can only be applied to specific scenes.

In summary, masked face detection methods based on AdaBoost algorithm and haar-like features are typical conventional methods. They can work well for close-distance scenes that have evident features of face regions.  However, due to limited learning ability, it is hard for these classifiers to adapt to complex scenes such as long distance and illumination changes.   Neural network-based methods are data-driven framework that may provide feasible solutions.

\subsection{Single-stage (End-to-end) Methods}

Single-stage methods based on deep learning techniques account for the largest proportion among the methods.  They include Faster R-CNN \cite{razavi2021automatic,meivel2021realMatlab}, Context-Attention R-CNN \cite{zhang2021novel}, InceptionV3 \cite{chowdary2020face}, MobileNet \cite{dey2021mobilenet}, SSD \cite{deng2021improved},  YOLO \cite{wang2021wearmask}, YOLOv2 \cite{loey2021fightings}, YOLOv3 \cite{jiang2021real,ren2020mask,prusty2021novel,dondo2021application,bhuiyan2020deep}, YOLOv4 \cite{kumar2021scaling,yu2021face,bhambani2020real,degadwala2021yolo}, YOLOv5 \cite{sharma2020face,Ding2020real,yang2020face,ieamsaard2021deep},  and others  \cite{ge2017detecting,jiang2020retinamask,rahman2020automated,omar2021automated,pooja2021face,mohan2021tiny,xiao2020application,prasad2021maskedfacenet,jian2021face,boulos2021facial}, etc. It can be clearly concluded  that YOLO and its variants are used widely. Representative methods are presented as follows .  

\textbf{Faster R-CNN-based:} Razavi et al \cite{razavi2021automatic} employed Faster R-CNN structure to detect people who do not wear a mask or do not maintain a safety distance.  It was applied to several road maintenance projects for monitoring workers,  ensuring them wear masks and keep proper physical distance. However, the dataset is limited and it only focuses on construction scenes.  Meivel et al \cite{meivel2021realMatlab} used Faster R-CNN algorithm for mask detection and social distance measurement. This method achieves 93.4\% accuracy for complex scenes such as facial poses, beard faces, multiple mask types, and scarf images. Notably, the effects need improvement when converting surveillance images into bird-view images.

\textbf{Context-Attention R-CNN-based:} 
Zhang et al \cite{zhang2021novel} developed a new framework for masked facial detection called Contex-Attention R-CNN, which consists of multiple context feature extractor component, decoupling branches component, and attention component.  It is able to enlarge intra-class difference and reduce inter-class difference through extracting distinguishing features.  They also created a dataset that includes 8635 faces with different conditions for experimental verification.  The framework can achieve \(mAP=84.1\%\) on the given dataset, 6.8\% higher than that of Faster R-CNN with ResNet-50. However, the dataset is classes imbalanced.  

\textbf{InceptionV3-based:}
Chowdary et al \cite{chowdary2020face} exploited InceptionV3 pre-trained model to classify one whether wears a mask or not. The last layer of InceptionV3 is replaced by 5 layers, which is regarded as a transfer learning model.  It is reported to reach a 99.9\% on a simulated dataset. 

\textbf{MobileNet-based:} Dey et al \cite{dey2021mobilenet} proposed a MobileNetMask to prevent the transmission of SARS-COV-2, which is a deep learning method of multi-phase facial mask detection. The mask classifier depends on the ROI detection of SSD and ResNet-10. Due to the minimal processing capability and lightweight mobile-oriented model, MobileNet-V2 is a good selection  for  embedded systems. It is reported to achieve higher accuracy than other methods.  

\textbf{SSD-based:}  Deng et al \cite{deng2021improved} introduced attention mechanisms, inverse convolution and feature fusion to SSD structure for the task of wearing mask detection.  It achieves an \(mAP\) of 91.7\%, outperforming SSD with 85.4\% mAP. 
 
\textbf{YOLO-based:} Wang et al \cite{wang2021wearmask} proposed a holistic edge-computing framework to detect masked faces. It is a serverless in-browser solution by integrate YOLO, CNN inference computing, and WebAssembly techniques. This design minimizes extra devices. It has easy deployment, low computation costs, fast detection speed, and achieves \(mAP=89\%\). 
  
\textbf{YOLOv2-based:} Loey et al  \cite{loey2021fightings} developed a YOLOv2 with ResNet-50 detector for medical face mask detection.  The method includes two parts. The first is designed by deep transfer learning for feature extraction. The second part is implemented by YOLOv2 for masked face detection. Specially, mean IoU is introduced to estimate the best number of anchor boxes and it can improve the accuracy.  The method achieves  \(AP=81\%\)  on a dataset with 1415 images. 
  
\textbf{YOLOv3-based:} Jiang et al \cite{jiang2021real} designed Squeeze and Excitation(SE) YOLOv3 to balance the effectiveness and running speed for masked facial detection.  It introduces SE into Darknet-53 as attention mechanism integration to extract essential feature, and adopts GIoUloss, focal loss to enhance stability and robustness.  A new dataset called Properly Wearing Masked Face Detection(PWMFD) Dataset  is created for three categories of masked faces. It is reported that the method achieves  \(mAP= 73.7\%\) for \(608\times608\) size of images. The method is expected to used in access control gate system and non-contact temperature measurement.   However, the similarity between incorrect masks is high.  It may bring confusions that masks only covering chin are regarded as without mask.    Prusty et al \cite{prusty2021novel} proposed a data augmentation technique to expand dataset size. New dataset is used to train YOLOv3 model for masked facial detection. Average accuracy is more than 93\% on given three datasets. However, only two kinds of data augmentation techniques (grayscale and Gaussian blur) are used. The number is very limited.  
  
\textbf{YOLOv4-based:} Kumar et al \cite{kumar2021scaling} explored to test original and tiny variants of YOLO on a new face mask detection dataset which encompasses 52635 images.  For the dataset, over 50k labels are provided.  Modified tiny YOLOv4 is recommended as an effective and efficient masked face detector because of its optimized feature extraction network.     Yu et al \cite{yu2021face} improved YOLOv4 model by introducing a modified CSPDarkNet53 to reduce computation costs and enhance learning ability.  An adaptive image scaling algorithm is designed to reduce redundancy and an improved PANet structure is used to learn  more semantic information. It is reported to achieve  98.3\% accuracy with 54.57 fps under the running environment of Windows 10, Inter(R)i7-9700k and RTX 2070Super. One limitation is inconsideration of insufficient lighting samples. 
  
\textbf{YOLOv5-based:} Sharma \cite{sharma2020face} developed a model that uses YOLOv5 to detect whether one person is wearing a mask or not. However, if an individual does not face the camera, its performance will decrease. This is the method's limitation.   Yang et al \cite{yang2020face} applied YOLOv5 in the supervision of wearing mask conditions. The authors design a man-machine interface for application and set the identifying time for 2 seconds with the consideration of complex scenes.  A 97.9\% recognition rate is achieved on the dataset  \cite{AIZOOTech}. It seems the response time is a bit longer.   Ieamsaard et al  \cite{ieamsaard2021deep} tested the performance of YOLOv5-based model with 300 epochs,  outperforming those models with less than 300 epochs.  
  

\textbf{RetinaFaceMask-based:}
Jiang et al \cite{jiang2020retinamask} proposed RetinaFaceMask for masked face detection, which is based on RetinaFace \cite{deng2020retinaface}. RetinaFaceMask is a single-stage detector. Its principle is to employ  feature pyramid network to fuse high-level semantic information. A novel context attention module is presented to help RetinaFaceMask focus on the features of faces and masks. Moreover, a cross-class removal algorithm is proposed to remove those regions with low scores and high IoU values. Experiments demonstrate that RetinaFaceMask outperforms RetinaFace \cite{deng2020retinaface} in \(Recall\) and \(Precision\).  

Moreover, there are more experimental comparisons between methods. Singh et al  \cite{singh2021face}  utilized two object detection models named Faster R-CNN and YOLOv3 for masked facial detection.  They presented the comparison from visual and quantitative views, and gave detailed discussions about the application. Faster R-CNN outperforms YOLOv3 in the accuracy,  however, for real-time application, it would be preferred to use YOLOv3 which runs faster than Faster R-CNN.  The selection of model depends on the environment conditions.  Similar conclusion is drawn in \cite{alganci2020comparative}.   Roy et al \cite{roy2020moxa} used SSD, Faster R-CNN, YOLOv3, and YOLOv3Tiny to cope with the challenges of wearing medical mask detection. These methods are tested on Moxa3K dataset. Experimental results demonstrate that YOLOv3Tiny is the most suitable method for real-time inference among the methods.

In summary,  object detectors such as Faster R-CNN and YOLO series attract more researchers' attentions, especially YOLOv3, YOLOv4 and YOLOv5. Tiny YOLO-based detectors with light-weighted models are expected to be deployed on real-time processing devices. Improved face detectors like RetinaFaceMask are also promising techniques. By transfer learning strategy, existing object detectors and face detectors can be applied for masked facial detection.

\subsection{Two-stage Methods}

Two-stage methods mainly encompass two stages: face pre-detection and face class verification.  The face pre-detection stage is usually implemented by many face detectors  \cite{zafeiriou2015survey,kumar2019face,zhang2016joint,wang2017face,deng2020retinaface,ShiqiYu,chen2018adversarial,tang2018pyramidbox} or object detectors  \cite{ren2016faster,liu2016ssd,redmon2016you,howard2017mobilenets,zhao2019object,bochkovskiy2020yolov4,li2019dsfd}, etc. Notably, object detectors can also provide feature descriptors for candidate faces in the first stage.  The second stage is designed by various classifiers or models  \cite{Chen2018Broad, chang2011libsvm, he2016deep, krizhevsky2012imagenet,chen2020efficient,arslan2021fine}. Its aim is to determine whether one wears a mask, correctly or incorrectly. The combination of object detector and classification model can realize masked face detection task.  

According to the used features in literatures, two-stage methods can be divided into three groups:  Neural Network + Neural Network  \cite{wang2021hybrid,hussain2021iot,nagrath2021ssdmnv2,mercaldo2021transfer,batagelj2021correctly,chavda2021multi,joshi2020deep,snyder2021thor}, Neural Network + Hand-crafted Feature  \cite{meivel2021realIoT,loey2021hybrid,buciu2020color,oumina2020control,zereen2021two}, Hand-crafted Feature + Neural Network  \cite{lin2016masked,faizah2021implementation,das2020covid,rudraraju2020face,malakar2021detection,adhinata2021deep,tomas2021incorrect}. 

 \textbf{Neural Network + Neural Network:}  


One representative example refers to the method \cite{wang2021hybrid}.  The first stage is designed  by a deep learning transfer model: Faster R-CNN \cite{ren2016faster,szegedy2016rethinking} and the second stage is designed by broad learning system (BLS) \cite{Chen2018Broad}. Input image is sent to the pre-detection stage. Then many candidate regions are generated and they are further classified by trained BLS model which can remove false positives and keep masked faces. Finally, detected results are generated with labels. To train pre-detection model, annotated dataset is required, which is created using a tool called ``LabelImg" \cite{labelImg}.  The extracted faces and masks can be used to create classification datasets that are problem-dependent, for example, with/without mask, correct/incorrect mask.

The pre-detection in Faster R-CNN structure mainly includes four steps: extract feature maps, generate proposals by Region Proposal Networks (RPN), obtain fixed dimension of feature map, and object classification and location regression. Faster R-CNN has advantages over SSD and YOLO in accuracy \cite{alganci2020comparative}.  The verification stage employs BLS, which is a flat neural network structure with a very high training efficiency \cite{Chen2018Broad} and many variants have been proposed  \cite{Chen2018Universal,liu2020stacked,chen2020random}.  In practice, when a BLS model can not learn a task well, one effective way is to add feature nodes that is called incremental learning. This ensures efficiency in training phase. It does not need to retrain from the scratch \cite{wang2021hybrid}.  The combination of Faster R-CNN and BLS are verified to be effective on WMD dataset \cite{wang2021hybrid}. It achieves 97.32\% accuracy for simple scene and 91.13\% for complex scene.  BLS can be as a good selection for classification when training efficiency and small size of model are required in applications.  Detailed descriptions can be found in supplementary materials.

 \textbf{Neural Network + Hand-crafted Feature:}
Loy et al \cite{loey2021hybrid} developed a hybrid method of deep learning and machine learning to detect facial mask. It includes two components (or stages): ResNet-50 is used as feature extractor, and SVM, decision tree, ensemble method are used as classification models. The authors claimed that SVM classifier achieves  testing accuracy of 99.49\% in SMFD dataset \cite{Prajnasb}, outperforming decision tree and ensemble method.

Similar with \cite{loey2021hybrid}, the methods \cite{buciu2020color,oumina2020control} also choose SVM as the classifier in the second stage. Buciu \cite{buciu2020color} took the ratio of color channels into account to discriminate mask and no-mask images. SSD is used to locate the positions of faces.  Then the lower part of face is considered to construct feature vector called color quotient feature, which will be classified by SVM model. A recognition rate of 97.25\% is obtained. However, this method is sensitive to mask types, which is its potential weakness.  Oumina et al  \cite{oumina2020control} presented several combinations of multiple CNNs and K-NN or SVM, and conducted experiments. It indicates that the combination of MobileNetV2 and SVM achieves the best performance among the combinations, 97.11\% accuracy. More tests for the approach should be conducted on bigger datasets. 
 
Zereen et al \cite{zereen2021two} developed a two-stage approach to detect masked face and monitor the rule violations. It is based on the extraction of facial landmark. It firstly determines whether the target wears a multi-color mask or not by MTCNN, and secondly it determines whether the target wears a skin-color mask or not.  The method aims to detect five types of facial images including no mask, beard and mustache, one-color-mask, multi-color mask and skin-color mask.  It achieves an accuracy of 97.13\% and overcomes the problem of various-color mask detection, especially differentiates wearing skin-colored masks. However, the use of several techniques needs more computation costs, and the setting of empirical thresholds limits its adaptation ability.  
 
 \begin{figure*}
	\begin{center}
		\begin{tabular}{c}
			\includegraphics[height=6.0cm,width=18.3cm]{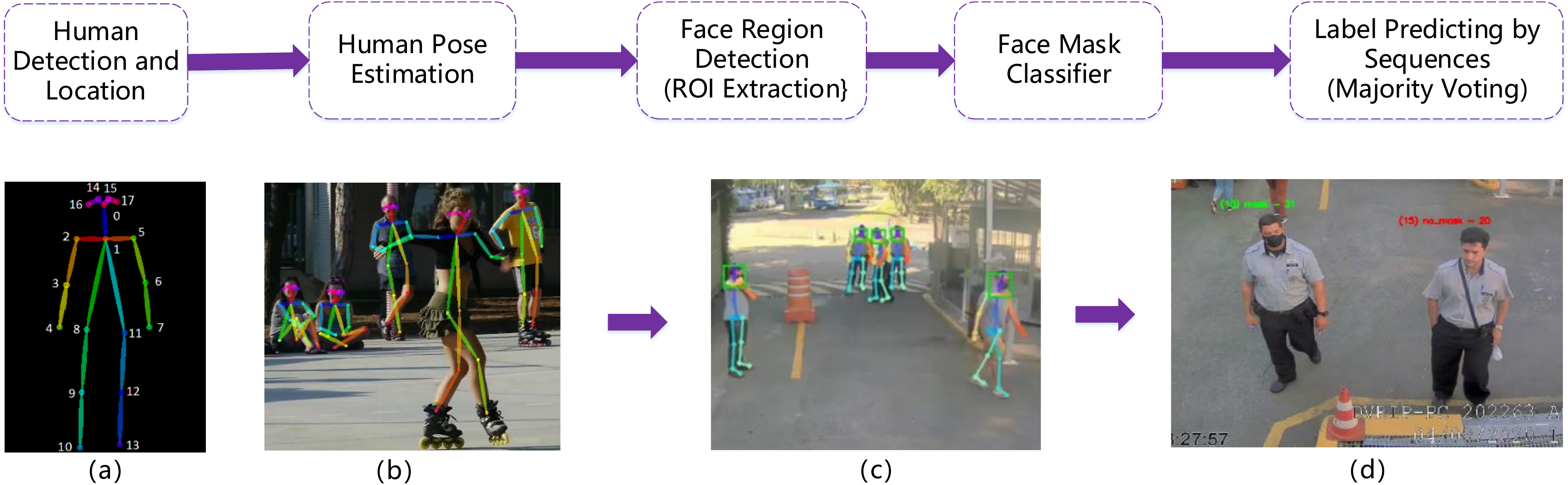}
		\end{tabular}
	\end{center}
	\caption 
	{ \label{fig:Multi-StageMethod}
		An example of multi-stage method for masked facial detection \cite{cota2020monitoring}. (a) Keypoints joints set, (b) 2D posture estimation, (3) Face region extraction, (d) Face/Masked face classification and labeling.  } 
\end{figure*}

 \textbf{Hand-crafted Feature + Neural Network:}  
Lin et al \cite{lin2016masked} combined a sliding window algorithm with a modified LeNet (MLeNet) to locate masked faces. To improve performance with a small dataset, horizontal reflection is used to learn MLeNet via fine-tuning. MLeNet can be trained fast under CPU mode. It makes sense for real-world applications. However, sliding window algorithm requires more computations for large size of images, which  restricts its performance. 

Rudraraju et al \cite{rudraraju2020face} combined haar-like cascade-classifiers and two MobileNet models for face mask detection. Firstly, face regions are detected by haar-like cascade-classifier. The first MobileNet  model is used to classify masks and no masks.  The second MobileNet model is used to distinguish correct or incorrect wearing masks. Experiments show that the system achieves around \(Accuracy=90\%\). It is expected to be deployed at fog gateway. 

Tomas et al \cite{tomas2021incorrect} also chosen haar-like cascade classifier for rapid facial detection. CNN with transfer learning is used to determine whether one wears a mask or not. Multiple models are trained based on one dataset. VGG16 achieves the best performance with 0.834 accuracy, but its model size is also the largest. For deploying mobile device, MobileNetV2, with 0.812 accuracy, is selected as the classification model because it demands less computation costs and smaller storage. However, this method needs to be improved when detecting masked facials with alterations and sides.

In summary, most of two-stage methods are the combination of face detector and classification model. In many situations, pre-detection model and classification model are trained separately, which might require more time than those of single-stage methods. However, two-stage methods have advantages in coping with small object detection, multi-class classification, and cross classes removal. The combinations of ``Neural Network + Neural Network" and ``Hand-crafted Feature + Neural Network" are attached more importance, and they provide feasible solutions to solve real-world problems.

\begin{table*}[htbp]  
	\caption{ A Brief Summary For The Representative  Methods. In Third Column, `1' Means Face Mask; `2' Means Face With Mask and Face Without Mask; `3' Means Face Without Mask, Face With Correct Mask, Face With Incorrect Mask; `4' Means  Face Without Mask, Face With Correct Mask, Face With Incorrect Mask, and `Mask Area'. }  
	\begin{center}
		\centering	
		\renewcommand\arraystretch{1.25} 
		\begin{tabular}{|p{1.8cm}|p{1.5cm}|p{1.23cm}|p{3.0cm}|p{3.8cm}|p{3.8cm} | }
			\cline{1-6} 
			\textbf{Category}	&	\textbf{ Methods}     & \textbf{ Detection Classes } & \textbf{ Datasets } &\textbf{ Results } &   \textbf{ Experimental Environment and Runtime}  \\ \hline
			
			\multirow{4}*{ \textbf{Conventional} }   & Dewantara et al \cite{dewantara2020detecting} & 2& 1000 images, self-built & \(Accuracy=86.9\%\) &     Image size: $50\times50$ to $275\times275$,  25fps \\	\cline{2-6}		
			
			& Nieto et al \cite{nieto2015system}  & 2 & 677 test cases, self-built & \(Recall=95\%\)&  VGA resolution $640\times480$, 10fps \\ \cline{2-6}		
						
			&   Petrovic et al \cite{petrovic2020iot}   & 3  & Not provide the number  & \(Accuracy= 84\%-91\%\)  &   Intel i7 7700-HQ quad-core CPU  2.80 GHz with 16GB RAM, image size $320\times240$, 38.46fps   \\ \cline{2-6}		
			
			& Fang et al \cite{fang2021design}   &  2  &   6024 images, self-built & \(Precision= 96.5\%\)  &   PYNQ-Z2 SoC platform, image size $1280\times720$, 45.79fps    \\ \hline

			\multirow{12}*{ \textbf{Single-stage} }   &		  
			Razavi et al \cite{razavi2021automatic} & 3   & 1853 images, self-built   &  \(Accuracy=99.8\%\)&   Not provide runtime  \\ \cline{2-6}	
			
			&Zhang et al \cite{zhang2021novel}  &  3  & 4672 images, self-built   & \(mAP=84.1\%\)&    Geforce GTX TitanX with memory 12G, not provide runtime   \\ \cline{2-6}	
			
			& Chowdary et al \cite{chowdary2020face}  &  2  & 1570 images,  simulated from SMFD \cite{Prajnasb} & Train \(Accuracy=99.9\%\), Test \(Accuracy=100\%\)&  Google Colab, not provide runtime  \\ \cline{2-6}	
			
			&Dey et al \cite{dey2021mobilenet}  &  2  &  3835 real images(IDS1), 1376 simulated images (IDS2)   & IDS1 \(Accuracy=93\%\), IDS2 \(Accuracy= 100\%\) &  Google Colab, not provide runtime   \\ \cline{2-6}	
			
			&Deng et al \cite{deng2021improved}   &  2  &  3656 images,self-built  & \(mAP=91.7\%\) &     NVDIA GTX 1070Ti GPU,  not provide  runtime \\ \cline{2-6}	 
			
			&Wang et al \cite{wang2021wearmask}   &  2  &  9097 images,self-built  &  \(mAP=89\%\)&  Google Colab (Tesla V100-SXM2-16GB),  not provide  runtime \\ \cline{2-6}	
			
			&	Loey et al  \cite{loey2021fightings}  & 1   & 1415 images, Kaggle \cite{Kaggle}    &  \(AP=81\%\) &  Not provide  runtime  \\ \cline{2-6}	 
			
			& Jiang et al \cite{jiang2021real} & 3   &  9205 images, self-built  & \(mAP=73.7\%\) &   RTX 2070 GPU with 8 GB memory, image size: $608\times608$, 64.0ms per image  \\ \cline{2-6}	 
				
			& Kumar et al \cite{kumar2021scaling} & 4   & 52635 images, self-built   &  \(mAP=71.69\%\) &   NVIDIA 1050i GPU with 8 GB memory, not provide  runtime \\ \cline{2-6}	
			
			&   Yu et al \cite{yu2021face} &   3   &   10855 images created from RMFD \cite{wang2020masked} and MaskedFace-Net \cite{cabani2020maskedface} &  \(mAP=98.3 \%\)  &  Inter(R)i7-9700k and RTX 2070Super with 8G memory, image size $416\times416$, 54.57fps \\ \cline{2-6}
			
			& Sharma \cite{sharma2020face}  &   2 &  Not provide the number  &  \(mAP\approx 60\%\) &   Not provide  runtime  \\ \cline{2-6}	
			
			& Jiang et al \cite{jiang2020retinamask}  &  2  & 7950 images, AIZOOTech \cite{AIZOOTech}    &  Face \(F1=93.73\%\), Masked Face \(F1=93.95\%\) &   NVIDIA GeForce RTX 2080 Ti, not provide runtime  \\  \hline

			\multirow{5}*{\textbf{Two-stage} }   & Wang et al \cite{wang2021hybrid}  & 1& 7804 images, WMD \cite{wang2021hybrid} & \(F1=94.19\%\) &  NVIDIA Geforce GTX 1660 super, 112.5ms per image   \\	\cline{2-6}	
			
			& Mercaldo et al \cite{mercaldo2021transfer}  &   2  &   4095 images from \cite{wang2020masked,Kaggle}  &    \(Accuracy=98\%\)   &   Intel Core i7 8th gen, equipped with 2 GPU and 16G RAM,  4.7s per  image  \\ 	\cline{2-6}

			& Loey et al \cite{loey2021hybrid} & 2 & DS1 \cite{wang2020masked}, DS2 \cite{Prajnasb},LFW \cite{huang2014labeled} & DS1 \(Accuracy=99.64\%\), DS2 \(Accuracy=99.49\%\)   &   Intel Xeon processor 2 GHz, DS1 0.203s per image, DS2 0.031s per image   \\ 	\cline{2-6}

			& Zereen et al \cite{zereen2021two}  & 2 & 5504 images, self-built &  \(Accuracy=97.13\%\) &   Not provide  runtime   \\	\cline{2-6}		
			
			& Rudraraju et al \cite{rudraraju2020face} &3 & 1270 images, self-built& \(Accuracy=90\%\) &    Not provide  runtime   \\ \hline

			\multirow{5}*{\textbf{Multi-stage} }   
			&Cota et al \cite{cota2020monitoring}  & 2 &2270 images, self-built & \(mAP=85.92\%\) &    NVIDIA GeForce GTX 1650 Max-Q with 4G memory, image size $320\times320$, 15.7fps \\	\cline{2-6}	
			
			& Lin et al \cite{lin2021near} &2 & 992 images, self-built  & Daytime \(Accuracy=95.8\%\),
			Nighttime \(Accuracy=94.6\%\) &   Daytime 1.826s per image, Nighttime 1.791s per image   \\	\cline{2-6}	
			
			&Qin et al \cite{qin2020identifying}  & 3& 3835 images, self-built & \(Accuracy=98.7\%\) &   A i7 CPU and P600 GPU with 4 GB memory, 0.03s per image   \\ \cline{2-6}		 
			
		     & Talahua et al \cite{talahua2021facial}   & 2   &  13359 images, self-built  &  \(Accuracy=99.65\%\)  & Google Colab, image size $224\times224$ , 0.84s per image   \\	\cline{2-6}	\hline

		\end{tabular}
		\label{tab:MethodsSummary}
	\end{center}	
\end{table*}

\subsection{Multi-stage Methods} 

Multi-stage methods always consist of multiple processing steps  \cite{cota2020monitoring,bu2017cascade,lin2021near,qin2020identifying,draughon2020implementation,talahua2021facial,mohammed2021smart}. For example, human detection or face region detection, ROI extraction or feature vector extraction, normalization, classification or prediction by sequences and so on. Alternatively, multi-stage methods can be constructed by different combinations of those components.    

The main idea of methods \cite{cota2020monitoring,lin2021near} is based on human posture estimation. Firstly, a certain number of key points for one person are estimated. Then, some key points in face regions are analyzed to extract ROI from original image.  After that, the ROI is normalized and sent to a trained classifier to predict class.  In practice, some additional operations may be required to enhance performance.

Fig. \ref{fig:Multi-StageMethod} shows the process of the method \cite{cota2020monitoring}.  It mainly includes five stages:  
\begin{enumerate}
	\item[(1)] Human detection and location is implemented by YOLOv4 \cite{bochkovskiy2020yolov4}. YOLOv4 is able to generate a series of candidates with a good trade-off between speed and accuracy in the field of object detection.  
	
	\item[(2)] Human pose is estimated by HRNet \cite{sun2019high}. About 18 key points are generated for each individual. This is can be found in Fig. \ref{fig:Multi-StageMethod} (a) and (b). 
	
	\item[(3)] Face ROIs are determined by the points belonging to eyes and nose. Only those key points with higher confidence ($>$0.8) are selected to determine valid faces. Meanwhile, the size of valid faces is restricted by \(20\times20\).  Too small ROIs will be removed. 
	
	\item[(4)] With valid faces obtained, they are classified by a transfer learning model \(ResNet101\times1\) \cite{kolesnikov2019big}. The model is trained on a data augmentation dataset.  
	
	\item[(5)] For each person, it is assigned with an ID. DeepSort \cite{wojke2017simple} is used to store some statistics. For each frame, the predicted label will be inserted into a buffer when the label's score is higher than 0.8.  The final label is estimated from the buffer (size$>$3) by the most frequent label, i.e., majority voting.     	 
\end{enumerate}

YOLOv4 is trained on 1370 images containing face and masked face classes, and it achieves an \(mAP\) of 85.92\% (IoU=0.5) on nearly 900 validated images. However, it does not perform well (\(mAP=40.3\%\)) on images with small size and low resolution. For classification, \(ResNet101\times1\) reaches 99\% for both classes: face and masked face.    

The method proposed by Lin et al  \cite{lin2021near}  contains five stages: image data collection, human posture parsing, ROI selection, image normalization, and classification of masked face.  Among these stages, human posture parsing is implemented by Openpose \cite{cao2019openpose} that  generates 25 key points for one individual. Five key points belonging to face region are used to extract ROI for image normalization. Then, the normalized image is classified by a Face Mask Recognition Network(FMRN). It is reported that the method obtains 95.8\% and 94.6\% accuracy in daytime and nighttime, respectively.      

Unlike \cite{cota2020monitoring,lin2021near}, Qin et al \cite{qin2020identifying} proposed a multi-stage method including four steps: image pre-processing, face detection and cropping  \cite{zhang2016joint}, image super-resolution, and wearing condition identification of face mask.  The distinctiveness of this paper is the introduction of super-resolution  network (SRNet) in  \cite{qin2020identifying}. The goal of SRNet is to enhance face image. It helps improve the  accuracy of subsequent classification network of mask-wearing condition.  The method is claimed to achieve an accuracy of 98.7\%. With the use of SRNet, it outperforms conventional deep learning method without SRNet by 1.5\%. However, it needs many calculations when three networks are carried out.  Meeting the requirement of real-time processing  is still a challenging task.
 
 Muhanad Ramzi Mohammed \cite{mohammed2021smart} et al developed a smart surveillance system to monitor one's mask-wearing condition and respecting social distancing. It includes three stages: the first stage is to detect humans using YOLOv3-tiny \cite{redmon2018yolov3};  based on the regions of detected people, SSD with a ResNet is used to detect face regions; then, MobileNetV2 is used to determine one whether wears a mask or not; finally, the detected ones will be compared with identification database to finish the recognition process. Several networks are used in the process, which seems a complex framework. Although every network has good efficiency, it is inevitable to take more time to perform all the networks.  How to reduce inference time and retain the performance is a worthy of study.  

In summary, multi-stage methods mentioned above have at least two deep learning networks. The design of multiple stages is relative complex compared with one-stage and two-stage approaches. It primarily focuses on performance improvement of masked facial detection. Experimental results of original literatures also demonstrate this point.  The drawback is also evident: multiple networks require many computations and expensive processing devices such as GPU.

\subsection{Discussions on the Results of Methods}
Before the outbreak of COVID-19, very limited number of papers were proposed for masked facial detection \cite{ge2017detecting,nieto2015system}. One important reason is the lacking of masked face datasets. As one of occlusions, masks account for a low ratio in many face detection datasets.  The COVID-19 epidemics accelerate the creations of masked facial detection datasets and give a rise to the research of masked facial detection methods. 

This paper present a roughly categories for the masked face detection techniques according to the used features and the number of processing stages. An overview of some representative methods mentioned are listed in Table \ref{tab:MethodsSummary}. It can be concluded that most of these methods are tested on their own datasets. We try to analyze them from three parts:   

\begin{itemize}
	\item Detection Classes: One-class detection means that only masked face is the objective in image or video. Most of masked face techniques are designed for two-class detection or three-class detection. Two-class detection methods determine whether one wears a mask or not. Three-class detection methods aim to detect face without mask, face with correct mask, face with incorrect mask. The four-class detection covers ``mask area" class additionally. In real-world applications, two-stage methods or multi-stage methods always locate the face regions firstly, then determine the mask-wearing conditions by further classification. In contrast,  single-stage neural network methods are able to detect multiple classes through a forward pass process.    
	
	\item  Datasets and Their Sources:  For each family of approaches,  thousands of images with annotations are used for training and testing except  \cite{nieto2015system,lin2021near}.  This is common requirement for neural network-based methods  with  supervised learning.  Images or datasets used in  \cite{chowdary2020face,dey2021mobilenet,jiang2020retinamask,loey2021hybrid} are from some existing datasets. The rest of methods in the \ref{tab:MethodsSummary} make use of their self-built datasets.  In this paper, we survey a series of available datasets in Table \ref{tab:OpenDataset}.  Different datasets can be combined for researchers to meet requirements and help solve the classes imbalanced problem.  Additionally, simulating samples is an alternative way to enrich datasets. Diverse types of masks will make contribute to the performance improvement of models.    
		 
	\item Results: It is hard to evaluate the best performance for all methods in Table \ref{tab:MethodsSummary} because they are tested on different datasets. It remains a work to compare these methods on a uniform dataset. Existing results with \(mAP\) and \(Accuracy\) can be regarded as a reference. On the other hand, various scenes can measure the adaptability of algorithms, for example, daytime and nighttime in \cite{lin2021near}. In terms of current results, it is believed that the most promising  detectors will be neural network-based techniques due to their strong learning ability and adaptability to significant variations in appearance of masks.

    \item Experimental Environment and Runtime:  Efficiency is an important metric to measure one approach.  However, quite a number of methods do not provide detailed descriptions about efficiency in Table \ref{tab:MethodsSummary}. For example, no information about experimental environment and runtime is provided in methods \cite{razavi2021automatic, loey2021fightings,sharma2020face,zereen2021two,rudraraju2020face}. The literatures \cite{zhang2021novel, chowdary2020face, dey2021mobilenet, deng2021improved, wang2021wearmask, kumar2021scaling,jiang2020retinamask} only give their environmental environments or test platforms, without runtime.  One potential reason may be derived from that researchers attach  more importance  to the performance or accuracy. The rest of methods in Table \ref{tab:MethodsSummary} shed better light on the runtime.  Due to different running environments such as GPU types and various image sizes, it's inapplicable to give a fair comparison between methods. It's clearly shown in  Table \ref{tab:MethodsSummary} that some conventional methods like \cite{dewantara2020detecting, petrovic2020iot, fang2021design} achieve the real-time processing effects without GPU.  In contrast, some CNN-based methods  \cite{mercaldo2021transfer,lin2021near}  are time-consuming because of the operation of more than one networks. Neural network-based methods are expected to be optimized to reach real-time processing while maintaining high accuracy.

\end{itemize}

   \begin{figure*}[h] 
 	\begin{center}
 		\begin{tabular}{c}
 			\includegraphics[height=14.5cm,width=14.0cm]{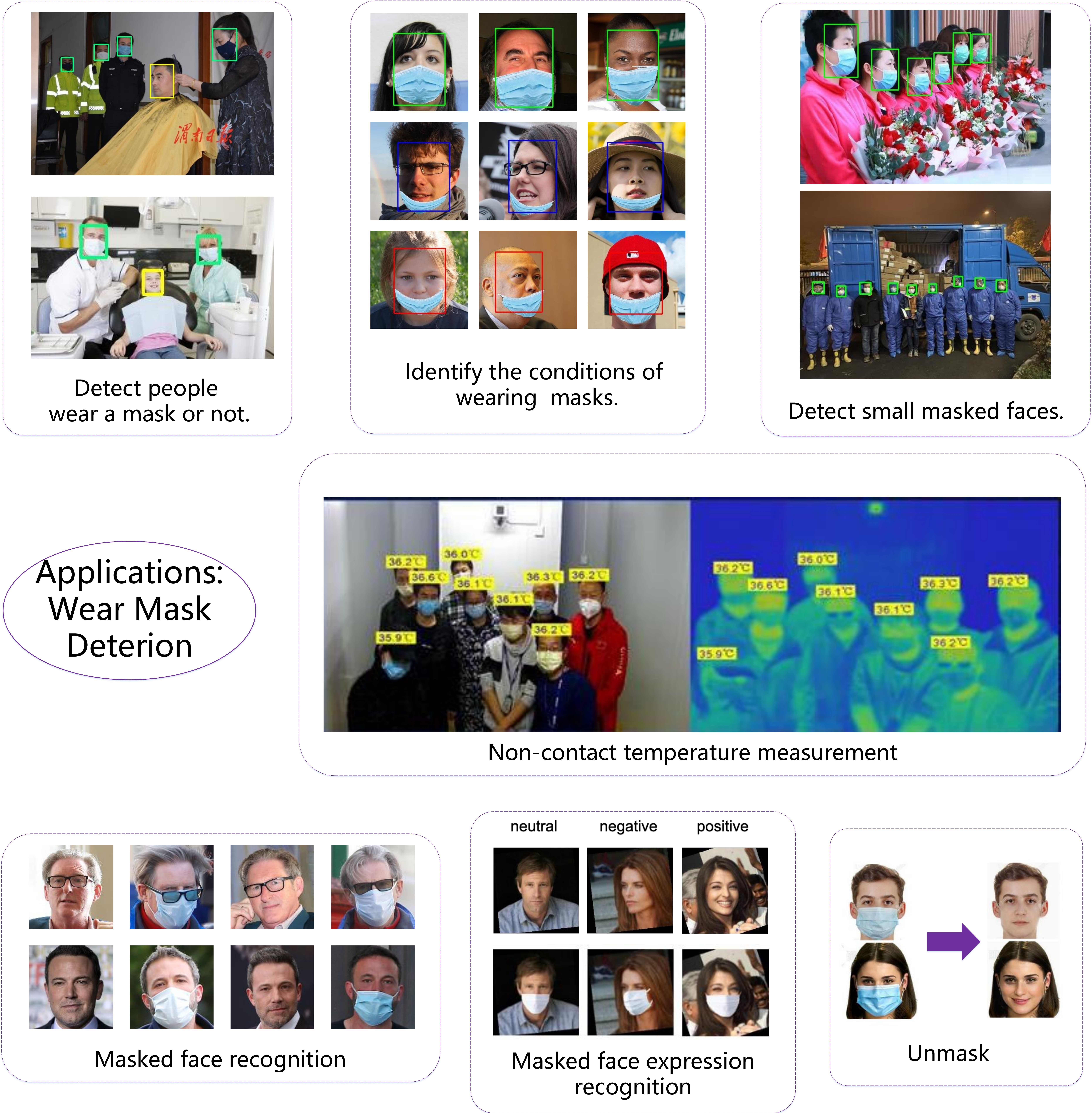}
 		\end{tabular}
 	\end{center}
 	\caption 
 	{ \label{fig:Applications}
 		Some applications of masked facial detection. The first row are generated by  \cite{wang2021hybrid}.  Non-contact temperature measurement is shown in second row, which comes from ``https://gongyi.gmw.cn/2020-03/23/content\_33675137.htm". The images in last row from left to right are collected from  \cite{anwar2020masked,yang2020facial,din2020novel}, respectively. } 
 \end{figure*} 
  
Moreover,  there are many applications based on masked facial detection methods in the era of COVID-19. Some examples are shown in Fig. \ref{fig:Applications}. Basic functions include: detect whether one wears a mask or not; identify the conditions of wearing a mask: correct or incorrect; detect small masked faces from long-distance views \cite{wang2021hybrid}. These functions are helpful for access control system, crowd counting, social distance monitoring, etc.  It should be mentioned that locating masked face regions can help infrared camera finish non-contact temperature measurement \cite{khan2020comparative} and reduce the infection risks caused by close-contact.  The results generated by masked facial detection methods can be sent to face recognition model to implement the identification verification \cite{anwar2020masked}.  Masked face expression recognition \cite{yang2020facial} is also an interesting application.  Masked faces can be used for unmask or face restoration \cite{din2020novel}, which is promising in the field of safety protection.

\section{Discussions and Future Research Directions}
 \label{section:DiscussionsAndFuture}
  	  
\subsection{Discussions on the Limitations of Datasets}
 
 Generally, one benchmark dataset is required with large quantity, multiple classes of wearing mask conditions,  versatile types of masked faces, proper ratio between realistic images and simulated images, and diverse scenes. However, there are some limitations for current datasets. 
 Herein, we discuss five points about the limitations of datasets.
 \begin{itemize}
 	\item Some datasets are (very) small in quantity.  For example, only several hundreds of images in the datasets are used for training deep learning models, which easily results in overfitting phenomenon.  In most cases, the larger the dataset, the better the trained model. 
 	
 	\item A fair proportion of datasets include only two classes: mask and non-mask. These datasets are only designed for distinguishing masked face from non-mask face.  Although some datasets include correct wearing mask and incorrect wearing mask, the number of incorrect wearing masks is very small.  
 	
 	\item Some datasets are created by simulating masks. Their quantities are always large. It makes for the training of masked face detection approaches. However, the mask type is always unitary when simulated.  Versatile types of masked faces are required to enrich those datasets. 
 	
 	\item Realistic and simulated images are both included in some datasets. However, the ratios  between realistic masks and simulated masks are imbalanced. The resolutions of images in some datasets may be in varied forms.     
 	
 	\item Most of images in some datasets are collected or captured from simple scenes. They are easily biased toward to a special scene. Thus, trained model based on such datasets may be ineffective for a new scene.  GAN-based techniques are expected to create various masked faces with different textures, colors and backgrounds.  

 \end{itemize}

\subsection{Discussions on the Limitations of Methods}

In the task of masked facial detection, there are some limitations for current methods. 
\begin{itemize}
	\item  Masked facial wearing conditions.  Some methods only detect two classes: masked facial or no-mask facial, ignoring of mask wearing conditions. It is well-known that incorrect wearing mask can not counteract the  spread of COVID-19.  Only a few methods were proposed to detect the mask-wearing conditions.  Thus,  more algorithms should be verified on the detection of masked facial wearing conditions. 
		
	\item Insufficiency of uniform evaluation for methods. Although some literatures present an evaluation of several methods, it still lacks of uniform evaluation for so many masked facial detection methods. Different methods may be implemented on different platforms. The results provided by original literatures only give readers conceptual comparisons.  It is not easy to give a fair judgement.
	
	\item Deficiency of computation cost.  Good performance is achieved by quite a number of methods. However, the cost effectiveness and running environment are not detailed for some methods.  In real applications, running time is an important measurement metric. Maintaining good performance with a light-weight equipment is a challenging task for existing techniques.   
	
	\item Lacking of model size. Many methods do not provide the size of trained models or the size of parameters.  Actually, this is an important issue for real-time processing on edge devices with limited storage. Light-weight models are supposed to be highlighted because they are in the hopes of deploying in mobile devices or edge devices. 
	
	\item Variation of image resolution. Some deep neural networks need a fixed size of images as input. However, input images are always with various resolutions. To meet the requirement of fixed size, these images are resized to prepare them for subsequent steps. This may bring about low image quality and  facial region distortion,  decreasing detection performance. 		
 \end{itemize}

\subsection{Future Research Directions}

In this section, we would like to highlight the future research directions.  Even though it was demonstrated recently that neural network-based methods have achieved excellent results, there are still some issues should be invested further. We conclude  ten  directions as follows. 
\begin{itemize}
	
	\item Create more balanced datasets. Classes imbalance problem exists as shown in Table \ref{tab:ClassesBalance}. Neural network-based methods are all appearance-based, which requires enough balanced data to train models. From the surveyed datasets,  we find that the number of incorrect wearing mask is very limited.  Thus, the category of images should be added significantly.  Collecting sufficient samples is a time-consuming and expensive task.  Two strategies can be taken into account. Firstly, simulating techniques of matching an mask to face can be used to create samples \cite{ZamhownWear-a-mask,hu2021covertheface}.  In this process, adding a variety of masked face types can enrich existing datasets.  Secondly,  GAN-based techniques can be used to produce a series of synthetic images that are very similar with real masks directly. Various environmental illuminations and head poses are expected to be generated.  In addition, data augmentation techniques \cite{prusty2021novel} can also be considered to add  more masked facial face orientations.  Hence, the mentioned techniques are all expected to make sense in the process of constructing more balanced datasets.

	\item  Apply transfer learning techniques to masked facial detection. In past decades, various object detectors are proposed and achieve excellent results like Faster R-CNN \cite{ren2016faster}, SSD \cite{liu2016ssd}, YOLO \cite{redmon2016you}, and MobileNet \cite{howard2017mobilenets}. These detectors are trained on multi-class detection datasets. They can also be used to detect masked faces by transfer learning techniques  \cite{niu2020decade}. Masked face detection and segmentation based on Mask R-CNN  \cite{lin2020face} can be also considered as a way. It is expected to realize more multi-class detectors in future. Advanced works of object detection can also be employed for the task of masked facial detection, for example, DEtection TRansformer (DETR) \cite{carion2020end}, anchor-free deep learning detectors CenterNet \cite{duan2019centernet} and CornerNet \cite{law2018cornernet}. In particular, how to implement knowledge transfer from current dataset to a special dataset is an interesting and promising direction.  

	\item Combine pre-detector and verification model  for masked facial detection.  Most of two-stage methods are the combinations of face detector and classification model  \cite{wang2021hybrid, loey2021hybrid, rudraraju2020face}. Pre-detection stage can be implemented by face detectors. Verification stage not only focuses on classification task but also solves some error detections like crosses classes problem.  The combinations between two stages are feasible. Conventional  models like AdaBoost cascade-classifiers can be combined with state-of-the-art CNN classification models for masked face detection. Multiple neural network models can be combined together to reach a high accuracy.  It is an interesting research direction to make a proper selection with a good trade-off between accuracy and efficiency.   
	 	 
	\item Consider contextual information for masked facial detection.  Masked face is one part of body and it is linked with other body parts. Some literatures such as multi-stage methods  \cite{cota2020monitoring,lin2021near} are designed to detect key points of body, e.g., 18 key points or 25 key points.  Based on the points belonging to eyes and nose, face ROI can be estimated.  Due to the occlusion of masks, the features are less in images that are captured from long distance \cite{wang2021hybrid,roy2020moxa}. Contextual information like key points can be utilized to improve the accuracy of small masked face detection. To our understanding, human pose estimation offers a powerful way and it is a very promising direction.  
	
	\item  Explore light-weight models and deploy them on mobile or edge devices.  A good light-weight model should be with fast inference and high accuracy.  It is of importance to integrate real-world masked facial detection system with Internet of Things. Moreover, the proposed light-weight neural networks in the published literatures need to be conducted on the same dataset and platform. Uniform evaluation of these methods can make readers a good understanding of every method's performance, and guide users to select a proper algorithm to meet their requirements. This is a valuable research direction.
	
	\item Process various resolutions of images. Some deep neural networks require a fixed size of images as input. In general, images with different resolutions need to be resized. Actually, resized images easily result in object distortion and information deficiency, which is a potential restriction. How to process various resolutions of images in a feasible manner is an important issue in future work.  
	
	\item  Masked face reconstruction. This is also called ``removing mask objects from facial images" \cite{din2020novel,li2020look,boutros2021unmasking}.  It is a challenge task because more than half of face region is occluded by mask and it is non-transparent.  To reach the goal of unmasking, two stages may be considered.  Firstly, mask regions need to be segmented very accurately. The second stage is to synthesize masked facial regions and it needs keep whole coherency of face structure.  GAN-based approaches are regarded to be effective because of its strong learning ability.  Therefore, it is an interesting issue to explore image editing techniques or object removal techniques to attain global coherency and restore deep missing regions.  This is of help for the tasks of masked face recognition  \cite{geng2020masked,deng2021mfcosface} and masked facial expression recognition  \cite{yang2020facial,mo2021mfed}. 
      
    \item  Masked face recognition. With the pandemic-driven continuous use of facial masks, it poses a huge challenge to conventional face recognition systems. This motivate researchers to develop a system that performs well with masked facials  \cite{geng2020masked,deng2021mfcosface,du2021towards,anwar2020masked}.  The requirement is more imperative than before. To solve the problem, two directions can be considered: the first is to recover masked regions for facial feature extraction; the second is to generate occlusion-robust feature from masked faces.  A competition of masked face recognition held at 2021 International Joint Conference on Biometrics (IJCB-MFR-2021) \cite{ijcb-mfr-2021} attracted many participators around the world to submit their solutions  \cite{zhu2021masked,boutros2021mfr,deng2021masked}. It is reported to collect the largest masked face recognition dataset.  In future, the deployability of innovative solutions proposed in IJCB-MFR-2021 will be considered to make sense in people's daily life.  It is encouraged to propose excellent algorithms for masked face recognition further.

    \item Masked faces and other biometrics for  multi-modal identification. In the era of COVID-19, people are required to wear a mask when entering public places.  Single face recognition technique may fail when one wears a mask.  Multi-modal biometrics can help a lot.  It is an interesting topic to combine masked facial with palm print, thumb, finger vein to construct multi-modal biometrics for object identification \cite{amin2021person}.  
    
     \item Masked face alignment.  The goal of face alignment algorithms is to predict the positions of facial landmark or pre-defined key points on faces.  When one wears a mask, much facial information is missing, which brings about huge challenge to existing face alignment algorithms. Although some researchers \cite{sha2021efficient,wen2021towards} have proposed a few solutions based on neural networks to tackle the problem, there are still many worthwhile works to improve the accuracy and reduce inference time.  It is believed to be a promising research direction. 
       	
\end{itemize}

\section{Conclusion}
\label{section:Conclusion} 
In this paper, we survey recent advances in the field of masked facial detection. Masked facial datasets are firstly reviewed. Thirteen open datasets are concluded from various aspects and their valid links are provided. We analyze these datasets from image sources, reality of images, classes imbalance, and experimental results. They can be used to create new larger datasets.   Simulating  wearing masks is an alternative way to generate samples to enrich existing datasets and improve the robustness of deep learning models.      

We review a series of masked face detection methods. They are classified as two categories: conventional methods and neural network-based methods.  Five typical conventional algorithms are outlined briefly. For neural network-based methods, they account for the largest ratio  and further classified as three classes according to the number of processing stages: single-stage methods, two-stage methods, and multi-stage methods. For each class, representative methods are described in detail and some typical techniques are introduced briefly.  Moreover, we summarize the results of representative methods according to the original literatures.  Limitations of datasets and methods are discussed.  Neural network-based methods are the mainstream and promising techniques. Finally, we highlight ten research directions about masked facial detection in the future.  Our work is finished in the era of epidemics in the hopes of providing some help in the fighting against COVID-19.

\section*{Acknowledgment}

We thank the authors of mentioned literatures for the sharings of their datasets.



\bibliographystyle{IEEEtran}
\bibliography{IEEEabrv,IEEEexample}

\begin{IEEEbiography}[{\includegraphics[width=1in,height=1.1in,clip,keepaspectratio]{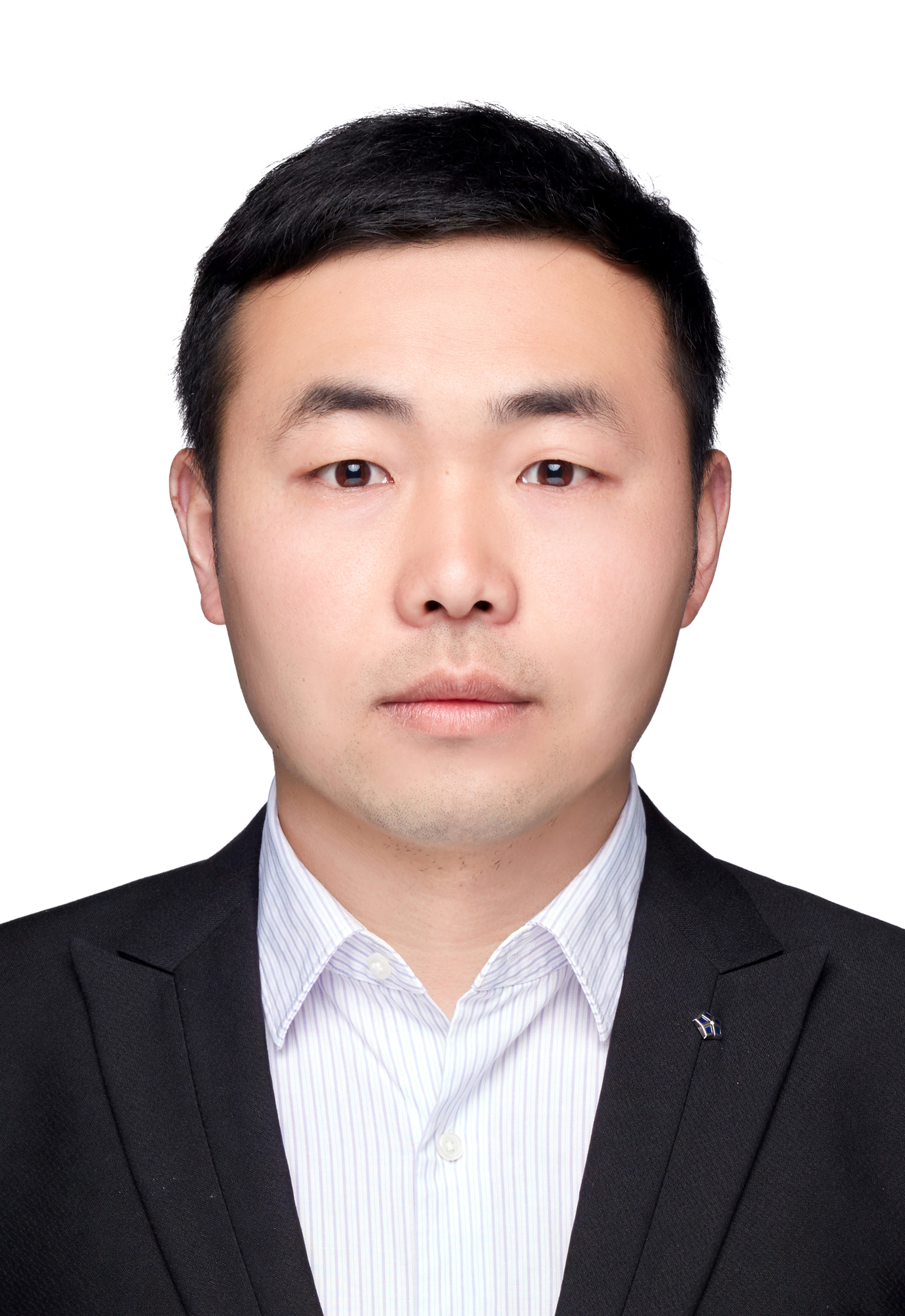}}]{Bingshu Wang} received his Ph.D. degree in Computer Science from University of Macau, Macau, China, in 2020. He received the M.S. degree in electronic science and technology (Integrated circuit system) from Peking University, Beijing, China, in 2016,  and B.S. degree in computer science and technology from Guizhou University, Guiyang, China, in 2013. Now he is an associate professor in School of Software, Northwestern Polytechnical University. He is also an member of Chinese Association of Automation (CAA). His current research interests include computer vision, intelligent video analysis and machine learning. 
\end{IEEEbiography}

\begin{IEEEbiography}[{\includegraphics[width=1in,height=1.1in,clip,keepaspectratio]{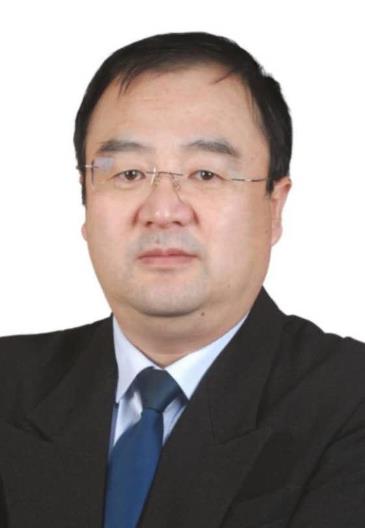}}]{Jiangbin Zheng} received the PhD degree from Northwestern Polytecnical University, in 2002, where he is a full professor and dean with the School of Software. His research interests include computer graphics, computer vision and multimedia.  He has published more than 100 papers in the above related research area. 
\end{IEEEbiography}

\begin{IEEEbiography}[{\includegraphics[width=1in,height=1.25in,clip,keepaspectratio]{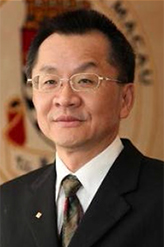}}]{C.L. Philip Chen}(S'88-M'88-SM'94-F'07) is the Chair Professor and Dean of the College of Computer Science and Engineering, South China University of Technology. Being a Program Evaluator of the Accreditation Board of Engineering and Technology Education (ABET) in the U.S., for computer engineering, electrical engineering, and software engineering programs, he successfully architects the University of Macau’s Engineering and Computer Science programs receiving accreditations from Washington/Seoul Accord through Hong Kong Institute of Engineers (HKIE), of which is considered as his utmost contribution in engineering/computer science education for Macau as the former Dean of the Faculty of Science and Technology. He is a Fellow of IEEE, AAAS, IAPR, CAA, and HKIE; a member of Academia Europaea (AE), European Academy of Sciences and Arts (EASA), and International Academy of Systems and Cybernetics Science (IASCYS). He received IEEE Norbert Wiener Award in 2018 for his contribution in systems and cybernetics, and machine learnings. He received two times best transactions paper award from IEEE Transactions on Neural Networks and Learning Systems for his papers in 2014 and 2018. He is also a highly cited researcher by Clarivate Analytics in 2018, 2019, and 2020.
	
Currently, he is the Editor-in-Chief of the IEEE Transactions on Cybernetics, and an Associate Editor of the IEEE Transactions on AI, and IEEE Transactions on Fuzzy Systems. His current research interests include cybernetics, systems, and computational intelligence. Dr. Chen was a recipient of the 2016 Outstanding Electrical and Computer Engineers Award from his alma mater, Purdue University (in 1988), after he graduated from the University of Michigan at Ann Arbor, Ann Arbor, MI, USA in 1985. He was the IEEE Systems, Man, and Cybernetics Society President from 2012 to 2013, the Editor-in-Chief of the IEEE Transactions on Systems, Man, and Cybernetics: Systems (2014-2019). He was the Chair of TC 9.1 Economic and Business Systems of International Federation of Automatic Control from 2015 to 2017.
\end{IEEEbiography}

\end{document}